\documentclass[a4paper,fleqn]{cas-sc}

\usepackage[authoryear]{natbib}
\usepackage{graphicx}
\usepackage{adjustbox}
\usepackage{todonotes}
\usepackage[T1]{fontenc}
\usepackage{verbatimbox}
\usepackage{graphics}
\usepackage{placeins}
\usepackage{supertabular}
\hypersetup{
    colorlinks=true,
    linkcolor=blue,
    filecolor=magenta,      
    urlcolor=blue,
}
\usepackage{xr-hyper}
\usepackage{hyperref}

\usepackage{cleveref}

\makeatletter
\newcommand*{\addFileDependency}[1]{% argument=file name and extension
  \typeout{(#1)}
  \@addtofilelist{#1}
  \IfFileExists{#1}{}{\typeout{No file #1.}}
}
\makeatother

\newcommand*{\myexternaldocument}[1]{%
    \externaldocument{#1}%
    \addFileDependency{#1.tex}%
    \addFileDependency{#1.aux}%
}

\myexternaldocument{supplementstuff}

\usepackage{comment}

\usepackage{gensymb}
\usepackage{booktabs}
\usepackage{adjustbox}
\usepackage{lscape}
\usepackage{subcaption}
\usepackage{caption}
\usepackage{multirow}
\usepackage{algorithm}
\usepackage{algpseudocode}
\usepackage{booktabs}
\usepackage{amssymb}
\usepackage[utf8]{inputenc}
\usepackage{color}
\usepackage{amsmath}
\usepackage{lineno}
\usepackage{setspace}
\usepackage[para,online,flushleft]{threeparttable}
\usepackage{makecell}
\def\tsc#1{\csdef{#1}{\textsc{\lowercase{#1}}\xspace}}

\tsc{WGM}
\tsc{QE}
\tsc{EP}
\tsc{PMS}
\tsc{BEC}
\tsc{DE}
\begin{document}
\let\WriteBookmarks\relax
\def\floatpagepagefraction{1}
\def\textpagefraction{.001}
\shorttitle{Deep Reinforcement Learning for Fire Prevention}
\shortauthors{L. Murray et~al.}
%\begin{frontmatter}

\title [mode = title]{Advancing Forest Fire Prevention: Deep Reinforcement Learning for Effective Firebreak Placement}

\author[1]{Lucas Murray}[]
\ead{lucasmurrayh@gmail.com}
\credit{}

\author[1]{Tatiana Castillo}[]
\ead{tatiana.a.castillo@gmail.com}
\credit{}

\author[2,3]{Jaime Carrasco}[orcid=0000-0003-4123-4228]
\ead{jcarrascob@utem.cl}
\cormark[1]
\credit{}

\author[1,3]{Andr\'es Weintraub}[]
\ead{aweintra@dii.uchile.cl}
\credit{}

\author[1,3]{Richard Weber}[]
\ead{richard.weber@uchile.cl}
\credit{}

\author[4]{Isaac Mart\'in de Diego}[]
\ead{isaac.martin@urjc.es}
\credit{}

\author[5]{Jos\'e Ram\'on Gonz\'alez}[]
\ead{jr.gonzalez@ctfc.cat}
\credit{}

\author[5]{Jordi Garc\'ia-Gonzalo}[]
\ead{j.garcia@ctfc.cat}
\credit{}

\address[1]{University of Chile, Industrial Engineering Department, Santiago, Chile}
\address[2]{Departamento de Industria, Facultad de Ingenier\'ia, Universidad Tecnol\'ogica Metropolitana, Santiago, Chile}
\address[3]{Complex Engineering System Institute - ISCI, Santiago, Chile}
\address[4]{Data Science Lab. Rey Juan Carlos University, Madrid, Spain.}
\address[5]{Centre de Ci\`encia i Tecnologia Forestal de Catalunya. Solsona, Spain.}

\cortext[cor1]{Corresponding author}

%%% ABSTRACT
\begin{abstract}
Over the past decades, the increase in both frequency and intensity of large-scale wildfires due to climate change has emerged as a significant natural threat. The pressing need to design resilient landscapes capable of withstanding such disasters has become paramount, requiring the development of advanced decision-support tools. Existing methodologies, including Mixed Integer Programming, Stochastic Optimization, and Network Theory, have proven effective but are hindered by computational demands, limiting their applicability.

In response to this challenge, we propose using artificial intelligence techniques, specifically Deep Reinforcement Learning, to address the complex problem of firebreak placement in the landscape. We employ value-function based approaches like Deep Q-Learning, Double Deep Q-Learning, and Dueling Double Deep Q-Learning. Utilizing the Cell2Fire fire spread simulator combined with Convolutional Neural Networks, we have successfully implemented a computational agent capable of learning firebreak locations within a forest environment, achieving good results.

Furthermore, we incorporate a pre-training loop, initially teaching our agent to mimic a heuristic-based algorithm and observe that it consistently exceeds the performance of these solutions. Our findings underscore the immense potential of Deep Reinforcement Learning for operational research challenges, especially in fire prevention. Our approach demonstrates convergence with highly favorable results in problem instances as large as $40\times 40$ cells, marking a significant milestone in applying Reinforcement Learning to this critical issue.

To the best of our knowledge, this study represents a pioneering effort in using Reinforcement Learning to address the aforementioned problem, offering promising perspectives in fire prevention and landscape management.
\end{abstract}


\maketitle

\section{Introduction}
\label{S:Intro}

The connection between climate change and the rising risk of wildfires underscores the need to rethink our approach to living with fire and our environment. Worldwide, fires are triggered by a variety of causes such as lightning, volcanic activities, accidental sparks from rock falls, or human carelessness, as detailed in \cite{scott2013fire}. In Canada, from 1990 to 2016, a combination of lightning, human activities, and other unknown factors was responsible for 47\%, 49\%, and 4\% of forest fires, respectively, as reported by \cite{tymstra2020wildfire}. In light of recent global events, including the devastating Australian bushfires of 2019 and 2020, California's massive Dixie Fire in 2021, the Fort McMurray wildfire in 2016, and Canada's ongoing 2023 fire season — which is on track to be one of the most destructive in the nation's history — it has become evident that merely reactive measures are not enough. These incidents underscore the urgent need for more proactive and comprehensive strategies in fire management and environmental conservation. Furthermore, they highlight the importance of exploring novel technologies to address this serious and escalating crisis.

Among the diverse preventive strategies employed in forest landscapes, such as cutting, clearing, controlled burning, and thinning, the establishment of firebreaks is a key method \citep{North2015ReformFF,carrasco2023firebreak}. This involves identifying strategic areas and replacing the existing vegetation with non-flammable materials. Consequently, these firebreaks serve as barriers, impeding the spread of fires should they occur. By halting the fire's progress, firebreaks play a crucial role in prevention and management. The implementation of preventive strategies in wildfire management has been increasingly informed by Operations Research (OR), integrating the fire risk and uncertainty into its mathematical models. Specifically, stochastic Optimization (SO) has emerged as a pivotal tool in this context, effectively accommodating uncertainty through a range of probabilistic scenarios, as evidenced in studies like \cite{hoganson1987model,boychuk1996multistage,eriksson2006planning,garcia2016accounting}. The latter study, in particular, underscores SO's potential in addressing climate change-related issues. However, the effectiveness of these methods can be limited when dealing with numerous scenarios. Combinatorial techniques have also been employed to tackle forest fire challenges, as demonstrated by \cite{gonzalez2011integrating}. Despite these advances, current research still faces significant challenges in optimally integrating fire risk into spatially explicit plans. These challenges include overly simplistic landscape models, reliance on a limited set of fire simulations, and the use of basic fire spread simulators, as indicated in the works of \cite{bettinger2009,kim2009spatial,konoshima2008spatial,gonzalez2011integrating}. Broadly, a notable shortcoming of these techniques is their lack of adaptability in learning from past experiences. Consequently, when managing a new landscape, models must be developed from scratch, failing to capitalize on the insights and knowledge gained from previously solved scenarios.

The advent of machine learning (ML) techniques in wildfire analysis and management opens up new opportunities to tackle this complex issue. ML enables computers to autonomously learn from data and encompasses three key types: i) supervised learning, which uses pre-labeled data to develop predictive models through classification and regression; ii) unsupervised learning, employing unlabeled data, often using clustering for exploratory data analysis; iii) reinforcement learning (RL), which aims to create models that maximize cumulative rewards through a series of actions, diverging from traditional input-output based learning. Since the 1990s, ML has been instrumental in various wildfire science and management areas, including fire occurrence and risk assessment \citep{amatulli2006assessing, costafreda2018human}, detection, climate change impact studies, effect analysis \citep{sousa2020wildfire}, behavior prediction \citep{hodges2019wildland}, wildland-urban interface mapping \citep{carrasco2019advanced, miranda2020evidence}, and fuel management \citep{lauer2017spatial}. However, most studies have primarily utilized supervised and unsupervised Learning, with a few notable exceptions such as \citep{lauer2017spatial, large-scale_rl}, which employed RL.

In RL, an agent learns optimal actions by interacting with its environment, aiming to maximize rewards in a Markov Decision Process (MDP), where transition probabilities are learned rather than predefined \citep{sutton1998introduction}. This learning style is ideal for automated decision-making tasks in areas like robotics or policy optimization. MDPs, however, struggle with the challenges of modeling in intricate systems (modeling curse) and managing extensive state spaces and transition probability matrices (dimensionality curse). This has spurred research into methods that operate within simulators, bypassing the need to generate transition probabilities, crucial in MDPs. Notable algorithms in this domain include Q-Learning \citep{q-learning}, the SARSA algorithm \citep{sutton2018reinforcement}, deep Q-network \citep{DQN}, and the Policy Gradient (PG) algorithm \citep{williams1992simple}, collectively advancing the field of RL. 

Traditionally, OR in wildfire management has focused on developing exact mathematical models. These models often require simplifying assumptions to be manageable, such as adopting specific distributions for random variables in the system or using representative scenarios in Stochastic Optimization. While these assumptions lead to mathematically elegant solutions, they can sometimes overlook the complexities of real-life situations. A notable challenge has been optimizing without closed-form expressions for objective functions, which provide direct mathematical solutions \citep{gosavi2015simulation}. Recognizing these limitations, our study adopts a reinforcement learning (RL) approach, renowned for its technical capabilities and effectiveness in handling complex, dynamic problems. Specifically, we develop a framework based on an RL algorithm that interacts with a wildfire simulator, Cell2Fire \cite{pais2021cell2fire}. This approach allows for a more nuanced and adaptable model, better suited to the unpredictable nature of wildfire behavior.

A common flaw of OR models is that they necessitate a fresh computational effort for each new problem encountered, as if no prior problem had been solved. This contrasts sharply with the capabilities of ML models, especially Deep Learning, which excel in leveraging accumulated knowledge. This knowledge comes in the form of parameters, which can be used either as a starting point for the new learning process or directly as a feature extractor after the input has been transformed to fit the dimensions of the input layer \citep{Goodfellow-et-al-2016}. Notably, Deep Learning models possess the unique advantage of combining specific insights from distinct problem instances, thereby enhancing solution efficiency and promoting the recognition of underlying similarities \citep{representation-learning}. This dual capability of accelerating the problem-solving process and integrating diverse instance-specific knowledge marks a significant departure from the conventional approach of OR models.

This study makes a dual contribution to the field. Firstly, it pioneers the application of RL techniques for addressing the Firebreak Placement Problem (FPP) marking a significant departure from conventional OR methods and positioning RL as a viable and promising alternative. Secondly, it conducts an exhaustive comparison of various algorithms, demonstrating their effectiveness and appropriateness for use in this specific context. To the best of our knowledge, this is the first instance of such an innovative approach being applied to the firebreak placement challenge.

\section{Materials and Methods}
\label{S:M&M}

\subsection{Firebreak Placement Problem in Perspective}
In this subsection, we define the FPP, building upon the groundwork laid by \cite{paper_david}. We consider a landscape segmented into a set of spatially georeferenced cells in the landscape, represented by $N$. Each cell is linked to a decision variable $x_{i}, i \in N$, which is assigned a value of 1 if a firebreak is placed in cell $i$ and 0 otherwise. This assignment results in a solution vector $x \in \{0,1\}^{|N|}$. Implementing a firebreak in a cell completely removes its vegetative fuel, rendering the cell non-combustible. Given the financial implications of firebreak placement, it is practical to limit the number of cells managed in this manner. In our approach, this limit is precisely $5\%$ of the total cell count in a forest. Formally, this constraint is expressed as:

\begin{equation} \label{Eq:Constraint}
\sum_{i \in N} x_{i} \leq \alpha |N| 
\end{equation}

where $\alpha \in (0,1)$ caps the overall quantity of firebreaks that can be deployed. The primary aim of placing firebreaks is to minimize the expected tally of cells consumed by fire under any arbitrary wildfire event $f$:
\begin{equation} \label{Eq:Objective}
\min_{x} \big(\mathbb{E}_{f}[L(x, f)] \big)
\end{equation}
In this expression, $L(x, f)$ denotes a stochastic variable representing the number of cells incinerated by a random fire event $f$ after firebreaks $x$ have been placed. In this work, $\mathbb{E}_{f}[L(x, f)]$ is unknown, relying instead on a series of simulated realizations of the variable $L(x, f)$ to model fire spread dynamics.

\subsection{Reinforcement Learning}\label{SS:RL}

\subsubsection{Basics of RL}
\label{SS:Basics}

RL is a ML paradigm in the intersection between supervised and unsupervised learning. In this, an agent interacts with an environment in a series of discrete time steps. In each time step $t \in \{1,2,..., T\}$, the environment presents an observation to the agent, generally known as state $s_{t} \in S$ until the episode ends at time $T$. In turn, the agent takes an action $a_{t} \in A_{s_{t}}$, where $A_{s_{t}}$ is the set of available actions at state $s_{t}$, causing the environment to transition to a new state $s_{t+1}$ and emit a reward signal $r_{t}$. The agent takes actions in order to maximize the sum of discounted rewards:
\begin{equation} \label{Eq:RL-Objective}
    G_{0} = \sum_{k=1}^{T}\gamma^{k-1}r_{k}
\end{equation}
where $\gamma$ is a discount factor that represents the notion that present rewards are more valuable than future rewards. The decision-making behavior of an agent is defined through the concept of a policy $\pi$, which is a mapping from states to actions. The full set of transitions an episode determines is called a trajectory: $\{(s_t, a_t, r_t, s_{t+1})\}_{t=0}^{T-1}$. Therefore, a successful resolution  of the RL problem implies determining the optimal policy $\pi^{*}$. A particular policy, determines what is known as the state-action function:
\begin{equation} \label{Eq:Q}
    q_{\pi}(s,a) = \mathbb{E}_{\pi}[\sum_{k=0}^{T}\gamma^{k}r_{t+k+1}|s_{t}=s,a_{t}=a]
\end{equation}
The former set of equations has a special formulation when it corresponds to the optimal policy called the Optimality Bellman Equations:

\begin{equation} \label{Eq:Bellman}
    q_{*}(s,a) = \sum_{(s',r)\in SxR}p(s',r|s,a)[r+\gamma \max_{a \in A} q_{*}(s',a')], \forall (s,a)
\end{equation}
where $p(s,a)$ are known as the transition probabilities, which enclose the whole environment dynamics and in general are not available. If not available, one has to recur to RL algorithms and in particular when the number of states is too large, to deep reinforcement learning (DRL) where it must be approximated: $ q(s,a,\theta) \approx q(s,a)$. Once $q_{*}(s,a)$ is found or approximated, the optimal policy will be to choose the action $a_{t}$ that maximizes $q(s_{t},a_{t},\theta)$ in each step. In order to ensure that the agent visits a variety of states, regardless of them being profitable, it takes a random action with probability $\varepsilon$. This type of policies are known as $\varepsilon$-greedy.

\subsubsection{Algorithms}
\label{SS:Algorithms}
As mentioned in Section \ref{SS:Basics}, the optimal policy in this work was learned through the approximation of $q_{*}(s,a)$, which in turn was performed through convolutional neural networks.

The general idea of the algorithms implemented is the following: the agent interacts with the environment for a number of episodes, collecting experiences $(s_{t}, a_{t}, r_{t}, s_{t+1})$ which are saved in a experience replay buffer $D$ (\cite{experience_replay}). Two copies of the Q-Network will be stored, one will be used to generate this experiences, denoted by $q(s,a,\theta)$. The other, $q(s,a,\theta^{-})$, will be used for computing the target value used in every step to update $q(s,a,\theta)$. After a $C$ number of episodes, the parameters $\theta$ will be copied into $\theta^{-}$. The rule used to update this parameters will be what is going to change between algorithms. 

The use of an experience replay buffer is a common practice in RL. For once, it provides greater sample efficiency than collecting and then discarding every time the network is updated. Secondly, neural networks assume i.i.d samples in each batch, something that is easier achieved if a large number of transitions are stored and then sampled from. Finally, it prevents the neural network from getting stuck in local optima by regularly providing it with transitions that were not generated by the current implicit policy. In a similar way, the use of a target network $q(s,a,\theta^{-})$ allows for learning processes considerably more stable than using a single network to both, collect experiences and generate target values.

To update the parameters the following loss function was used:
\begin{equation} \label{Eq:Loss}
    L(\theta) = \sum_{(s,a,r,s^{'})\sim U(B)}(target - q(s,a,\theta))^{2}
\end{equation}

with:
\begin{equation} \label{Eq:Target}
    target = r_{t} + \gamma max_{a' \in A}q(s_{t+1}, a', \theta^{-})
\end{equation}
where the ``target'' is an approximation of $q(s,a,\theta)$ based on Eq.\:(\ref{Eq:Target}) and therefore Eq.\:(\ref{Eq:Q}) should indicate how close is the estimate to the actual value. Eq.\:(\ref{Eq:Bellman}) gives place to the first algorithm, Deep Q-Learning introduced in \cite{DQN}. Although Deep Q-Learning in general works well, it suffers from a considerable flaw: it usually overestimates the values of $q_{*}(s,a)$, this is because if $q(s, a, \theta)$ contains errors, then the target as constructed in Eq.\:(\ref{Eq:Target}) is overestimated \citep{thrun2014issues}. One way to address this complication is to decouple action selection and evaluation:
\begin{equation} \label{Eq:Target2}
    target = r_{t} + \gamma q(s_{t+1}, argmax_{a' \in A}q(s_{t+1}, a', \theta), \theta^{-})
\end{equation}
This way of constructing the target defines the second algorithm: Double Deep Q-Learning (\cite{double_DQN}).

A final improvement to the presented algorithms introduces the third one: Dueling Double Deep Q-Learning (\cite{duelling_DQN}). In this, the network instead of outputting only $q(s,a,\theta)$, outputs $v(s,\theta)$ as well as $A(s,a,\theta)$. The first is known as the value-function and corresponds to the expected discounted rewards the agent obtains when faced with $s$ and then following a particular policy. The second is known as the advantage-function and similarly, it is the difference between the discounted rewards obtained by executing action $a$ in state $s$ in comparison to the expectation for all actions in that state. The benefits of estimating $v_{*}(s)$ and $A_{*}(s,a)$ instead of $q_{*}(s,a)$ directly is that in some states, it is not necessary to estimate each of the possible actions values, since they could have little effect in the environment. In such states, it is sufficient to use the value function, this leads usually to better performance. With these two functions $q(s,a,\theta)$ can be reconstructed and used as before:
\begin{equation} \label{Eq:DDQN}
    q(s,a,\theta) = v(s,\theta) + (A(s,a,\theta) - \frac{1}{|A|}\sum_{a^{'} \in A} A(s,a^{'}, \theta))
\end{equation}

Having this new formulation for $q(s,a,\theta)$, the target is constructed as in Eq.\:(\ref{Eq:Target2}). 

A particular advantage of these methods with respect to usual PG approaches is that there is no assumption about the distribution from which the experiences are sampled. In particular, one can generate experiences according to a completely different distribution with respect to the current policy, such as a demonstrator. This idea is known as learning from demonstrations \cite{DQN_demonstrations}. These demonstrations are stored in the replay buffer and will not be overwritten by experiences gathered by the agent. To incorporate these new experiences, new terms must be added to the loss function. Firstly, to ensure that the agent regularly chooses to follow the demonstrator behavior one forces those actions to have marginally larger values than the rest:
\begin{equation} \label{Eq:Loss-Expert}
    J_{e}(\theta) = max_{a \in A} [q(s,a,\theta) + l(a_{e}, a)] - q(s,a,\theta)
\end{equation}
where, $a_{e}$ is the action taken by the demonstrator when faced with state $s$ and $l(a_{e}, a)$ is a margin function that is 0 when $a=a_{e}$ and a positive value otherwise. In addition, to ensure that $q(s,a,\theta)$ is well adjusted for early steps of an episode, the n-step loss is included:

\begin{equation} \label{Eq:n-Loss}
    J_{n}(\theta) = r_t + \gamma r_{t+1} + ... + max_{a^{'} \in A} \gamma{n} q(s_{t+n}, a^{'},\theta)- q(s,a,\theta)
\end{equation}

And finally, to prevent parameters from getting too large, an L2 penalization is added:

\begin{equation} \label{Eq:L2-Loss}
    J_{L2}(\theta) = \sum_{w \in \theta} w^{2}
\end{equation}

Then, the global loss function is defined by summing up Eq.\:(\ref{Eq:Loss}), Eq.\:(\ref{Eq:Loss-Expert}), Eq.\:(\ref{Eq:n-Loss}) and Eq.\:(\ref{Eq:L2-Loss}):

\begin{equation} \label{Eq:Global-loss}
    L(\theta) = J_{DQ}(\theta) + \lambda_{1}J_{n}(\theta) + \lambda_{2} J_{e}(\theta) + \lambda_{3} J_{L2}(\theta)
\end{equation}
where in general, $\lambda_{1} = \lambda_{2} = \lambda_{3} = 1$. $J_{DQ}(\theta)$ corresponds to the loss function defined in Eq.\:(\ref{Eq:Loss}), where the target is calculated according to which of the three algorithms is to be used. Previous to the general training loop presented at the beginning of this section, a pre-training phase is added. In this phase, experiences are sampled from the experience replay buffer, which at this point is composed solely of demonstrations, and the network's parameters are updated according to Eq.\:(\ref{Eq:Global-loss}). This procedure allows the algorithm to learn how to mimic the decisions taken by the demonstrator, then this behavior is improved through the agent-environment interaction.

In addition to the aforementioned methods, others belonging to the PG family were initially considered, but then discarded because of the complexity of including demonstrations on their learning loops. Such is the case of Proximal Policy Optimization \citep{ppo} and Trust Region Policy Optimization \citep{trustpolicy}.

\subsection{Fire Growth Model}
\label{SS:FG-Model}
A crucial element in model-free algorithms is the ability to generate samples from the system within which the problem is framed. In the case of this work, that translates to being able to model the behavior of fire spreading and its interactions with the presence of firebreaks. In this study, we use a simulator known as Cell2Fire \citep{pais2021cell2fire}, which models fire spreading using a cellular automata approach, representing the landscape as a regular grid composed of cells characterized by a set of environmental variables, including fuel type and topographic features. Cell2Fire effectively integrates decision-making (e.g., firebreaks) with spatial simulation, an advantage that distinguishes it from other simulators such as Prometheus or Burn-P3 \citep{tymstra2010development,parisien2005mapping}.

\subsection{Modelling approach}
\label{SS:Modelling}
A crucial element for a RL system to work successfully is to feed the agent a suitable representation of the environment. This representation must contain enough information so that the agent can infer a considerable portion of the state of the system, in particular the immediate effects that its actions have on the environment. The representation designed in this work is based on \cite{vision_grids} idea of ``vision grids", which can be understood as a ``snapshots of the environment from the agent’s point of view".

\begin{figure}[h]
    \centering
    \includegraphics[width=0.8\textwidth]{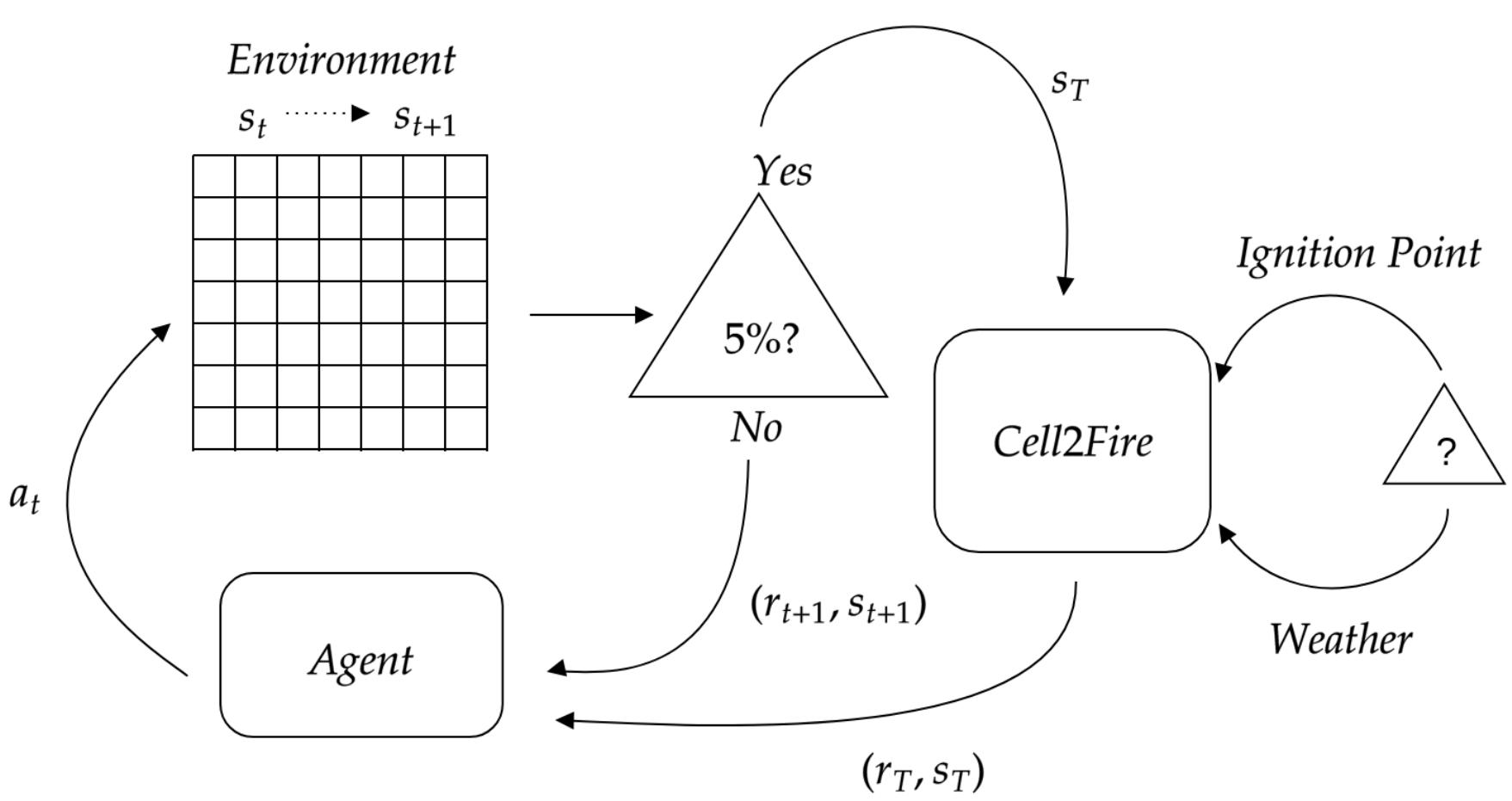}
    \caption{Agent-Environment interaction scheme.}
    \label{fig:agent-env}
\end{figure}

The general interaction loop consists of the agent choosing a single firebreak on each step of the episode. This will continue until the number of chosen firebreaks equals $5\%$ ($\alpha=0.05$ in Eq.\:(\ref{Eq:Constraint})) of the total number of cells in the forest, then a series of fires are simulated on Cell2Fire over the forest with the chosen firebreaks allocated. In order to conduct the simulation, a particular weather scenario and ignition point are sampled at random, determining the spreading direction and starting point. This is represented in Fig.\:\ref{fig:agent-env}.

Following the idea of vision grids, the agent is provided with a matrix representation of the environment of size $N\times N$. On each cell, there is a value that identifies uniquely the type of fuel present in it. 

\begin{figure}[h]
    \centering
    \includegraphics[width=0.6\textwidth]{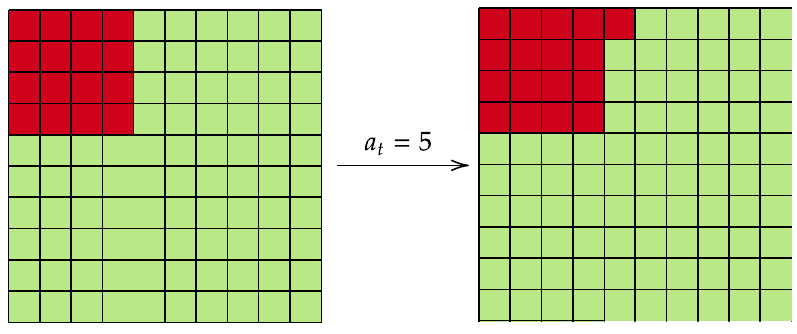}
    \caption{State transition when the fifth cell from the first row is chosen as a firebreak and therefore added to the forbidden set. The cells in red are in the forbidden set, the ones in green in the available.}
    \label{fig:s-a-representation}
\end{figure}

On every time step, there is a set of available and forbidden cells. This last set allows for the inclusion of restrictions of which cells can be chosen to be treated as firebreaks. Then, the agent chooses an action that corresponds to one of the cells in the available set. When a cell is chosen, the corresponding fuel value is changed for that of a non-fuel material and is added to the forbidden set (Fig.\:\ref{fig:s-a-representation}).

\subsubsection{Reward Function}
\label{SS:Rewards}
Another fundamental element in the construction of an RL system is to give the agent appropriate rewards for its actions. Specifically, maximizing this reward must be equivalent to solving the problem dealt with. A simple reward structure was used during this work:

\begin{equation} \label{Eq:Rewards}
    r(s_{t}, a_{t}) = \begin{cases}
        0 & t < T-1 \\
        nb(s_{t})\cdot k & t = T-1
    \end{cases}   
\end{equation}

where $nb(s_{t})$ is equal to the average number of cells that where burned on the simulator and $k$ is a penalization factor.

\subsection{Network Architectures}
\label{SS:Networks}
As was briefly discussed in previous sections, due to the number of states being too large to store in a table, function approximators must be used. "In particular, Convolutional Neural Networks were selected because of their well-studied success in visual recognition tasks.

Since $q(s,a)$ is going to be approximated, the output of the neural network is different than in most applications. The general intuition is the following: a particular state $s$ is fed to the neural network in the form of a (N,N) matrix. A series of convolution + max pool + dropout blocks are applied to the input, which act as feature extractors. In the case of the first two algorithms presented in \ref{SS:Algorithms}, these features are passed through one set of feedforward-fully-connected layers which act as a multi-output regression, outputting all the entries of $q(s,a)$. In the case of the third algorithm, another independent set of forward-fully-connected layers is added that perform a simple regression over the features, resulting in the state's value $v(s)$. The other set results in all the entries of the advantage function $A(s,a)$. In all three algorithms, entries corresponding to forbidden actions are masked.

\begin{figure}[hbt!]
    \centering
    \begin{minipage}[b]{0.4\textwidth} 
        \centering
        \includegraphics[width=\textwidth]{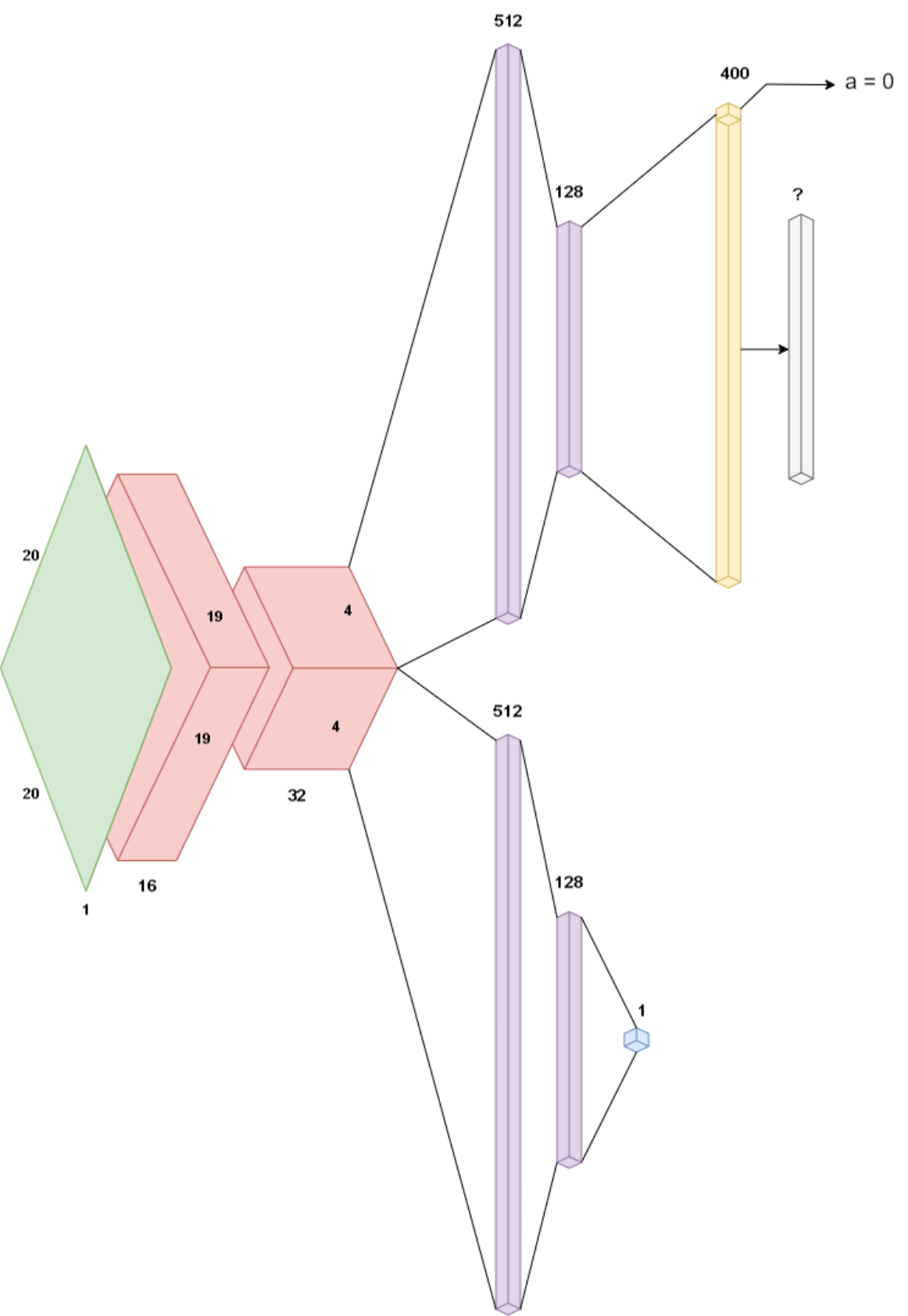}
        \subcaption[]{}
        \label{fig:small-net}
        \end{minipage}
    \begin{minipage}[b]{0.4\textwidth} 
        \centering
        \includegraphics[width=\textwidth]{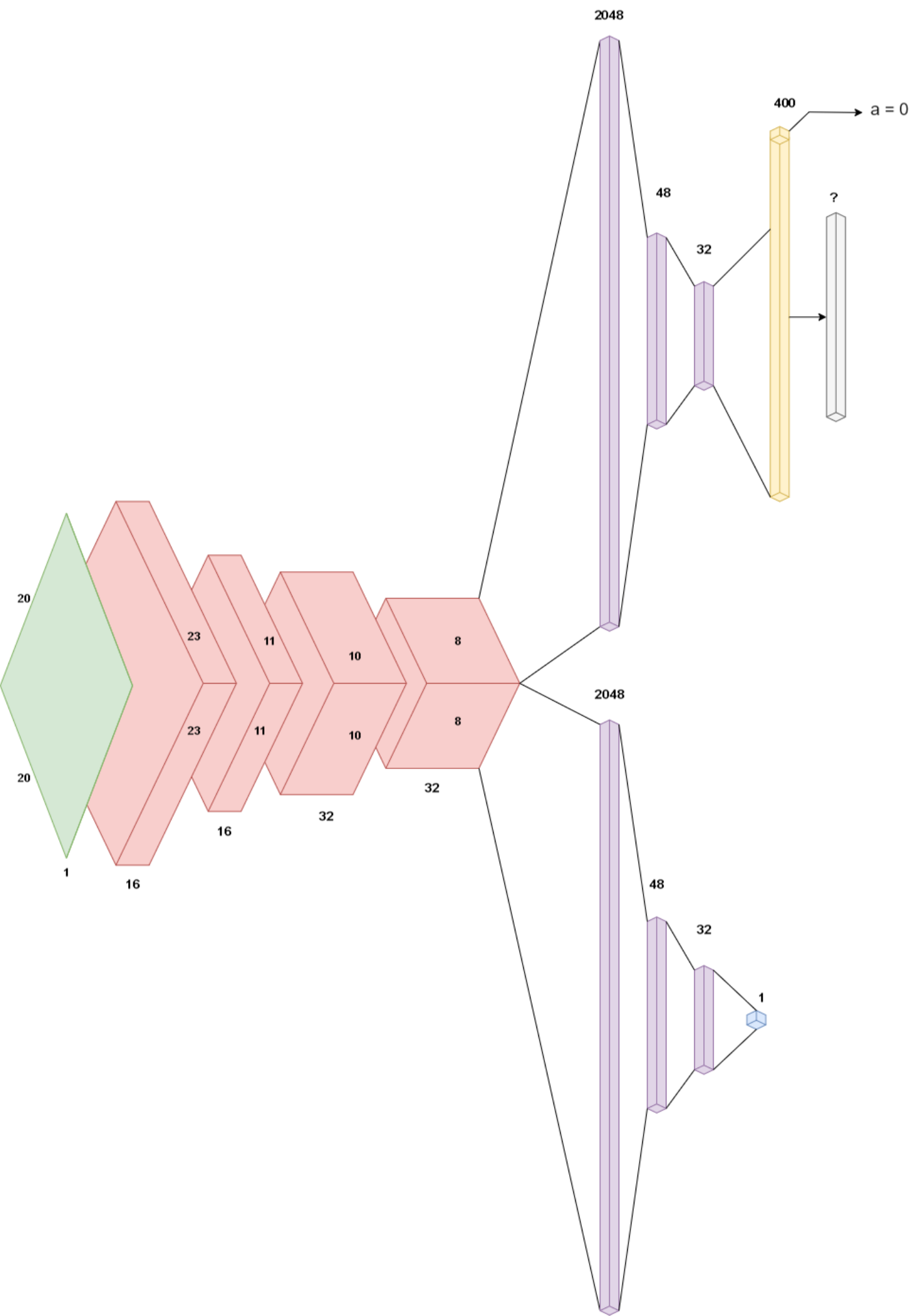}
        \vspace{0.25cm}
        \subcaption[]{}
        \label{fig:big-net}
    \end{minipage}
    \caption{small-net and big-net architectures, denoted by a) and b) correspondingly. Red blocks are convolution + max pool + dropout operations, purple bars are feedforward-fully-connected layers, the yellow bar is a mask, the blue cube corresponds to $v(s)$ and the grey bar is $q(s,a)$ or $A(s,a)$ depending on the algorithm.}
    \label{fig:networks}
\end{figure}

Two network architectures were trained, varying in their depth, in order to test how much complexity was needed to approximate $q(s,a)$. The first one, called small-net (Fig.\:\ref{fig:small-net}), is composed of two convolution + max pool + dropout blocks plus two separate output flows each consisting in two feedforward-fully-connected layers of 512 and 128 neurons correspondingly.

The second architecture, named big-net (Fig.\:\ref{fig:big-net}) has three convolution + max pool + dropout blocks and three feedforward-fully-connected layers per head of 2048, 48 and 32 neurons.  

Besides the proposed architectures, two larger pre-trained models were used, through a technique called Transfer Learning.

\subsubsection{Transfer Learning}
\label{SS:Transfer-Learning}
    Transfer learning is a ML technique that was introduced by \cite{transfer_learning}, yet its fruits only blossomed in the second decade of this century.Through this method, pre-trained, usually large models, are reused in order to construct a different model thought for a different yet similar task. The intuition behind it rests on the assumption that if both tasks are sufficiently similar, then some of the features extracted by the larger model should be similar to the ones the second task requires. Whether this premise holds is usually empirically checked. Then, the pre-trained model is used as a feature extractor for the second model, which generally adds a series of layers before and after the extractor. The point of this technique is to lighten the computational load for the second model, which might be critical if the data or resources are scarce. For the procedure to fulfill this, an arbitrary number of parameters in the pre-trained model are not adjusted or are ``frozen". 
    
    In this work, two pre-trained models were incorporated as feature extractors:
    \begin{enumerate}
        \item[i)] MobileNet: presented in \cite{mobile-net}, it is a model targeted for low resource applications, which translates into a reasonably small amount of parameters while not compromising performance. Two versions were used, MobileNetV3-Large and MobileNetV3-Small, this second having even fewer parameters than the first.
        \item[ii)] EfficientNet: presented in \cite{efficient-net}, it follows the same principle as MobileNet, yet focuses on balancing network depth, width, and resolution.
    \end{enumerate}

    Both models are considered state-of-the-art under resource constraints. To provide more details, in this application, a convolutional layer was added before the pre-trained models and a regressor after, varying slightly on their dimensions to match the size of the feature vector provided by each architecture. Regarding the parameters, the pre-trained weights were used as starting point and those after the 8th layer were ``frozen".

%\iffalse
\subsubsection{Explainability}
\label{SS:Explainability}
Most complex AI models are often referred to as ``black boxes" in the sense that, for most applications,
the results are not really understood or interpreted to determine which features of the input allow the
task to be successfully completed. To use these models lightly without any study of their properties is
no minor gamble. It is of particular importance in the context of this work, where decisions taken by the model translate into changes that affect individuals and communities and hence must be explained to some degree. 

\begin{figure}[pos=h]
    \centering
    \includegraphics[width=0.6\textwidth]{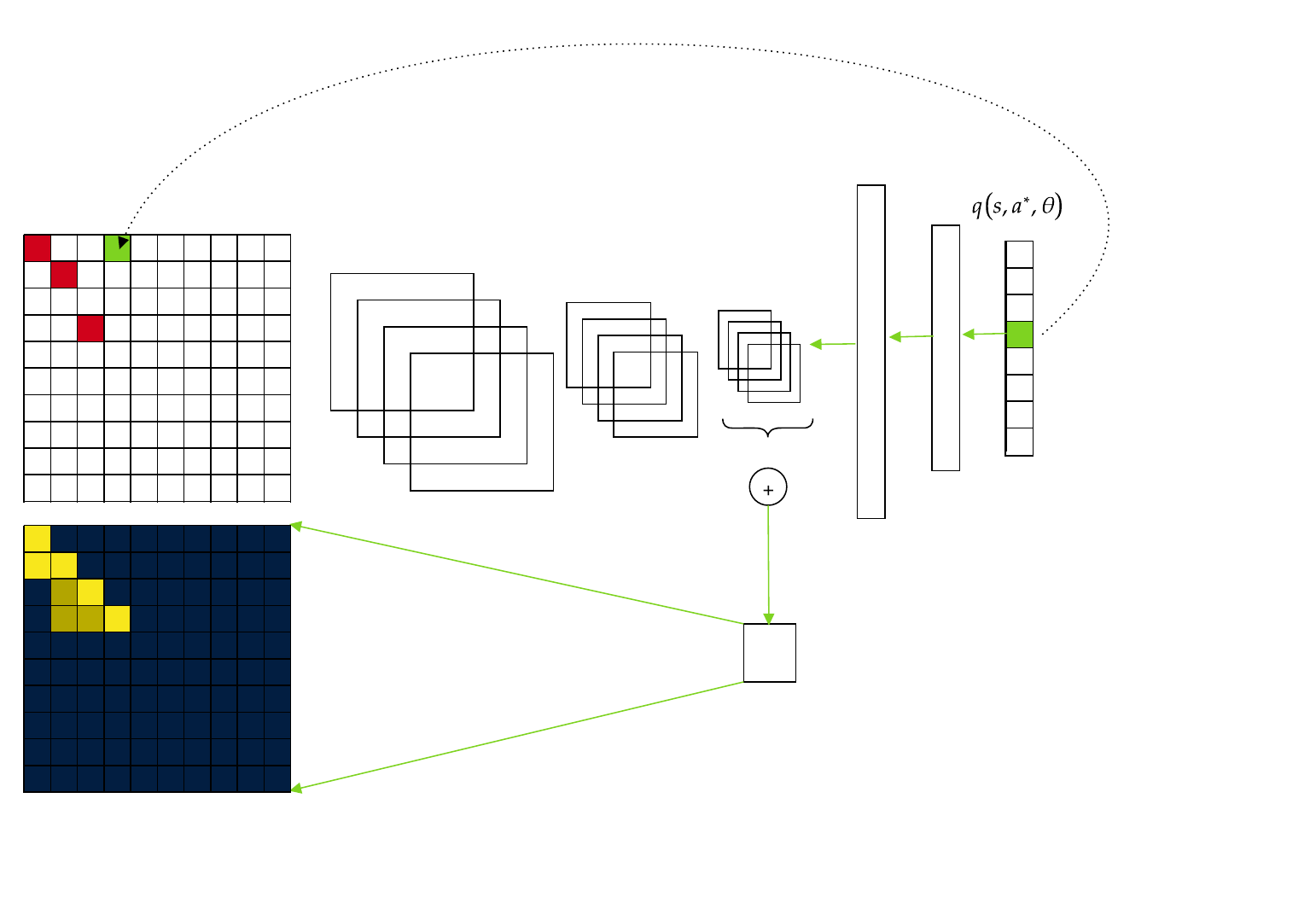}
    \caption{Gradient-weighted class activation procedure. The green cell of the output vector corresponds to the selected action. The computed gradients are combined in a single matrix, below the + operator and then scaled, generating the full attention map in blue-yellow below the input.}
    \label{fig:gradcam}
\end{figure}

In this work, following the work in \cite{pais2021deep}, we implemented a very common methodology to achieve certain level of explainability regarding the outputs generated by the model: gradient-weighted class activation or GradCam \citep{gradcam}. This method is widely used in computer vision and its extension to RL methods is straight-forward  according to \cite{interpretation}: given an action $a^{*}$ the gradient of the corresponding output entry $q(s,a^{*},\theta)$ with respect to the last convolutional layer of the network is computed. This gradient is a tensor which is combined and scaled in order to generate an image of the same dimensions as the original input, where larger values indicate that in order to compute $q(s,a^{*},\theta)$, the network puts higher relative importance to that pixel. A schema of this procedure is presented in Fig.\:\ref{fig:gradcam}.
%\iffalse
%\fi
\subsection{Experimental Setup}
\label{SS:Experimental-Setup}
\iffalse
\begin{figure}[pos=h]
    \centering
    \includegraphics[width=0.5\textwidth]{interpolation.pdf}
    \caption{Shrinking process through Nearest-neighbor Interpolation.}
    \label{fig:enter-label}
\end{figure}
\fi

\begin{figure}[h!]
    \centering
    \begin{minipage}[b]{0.53\textwidth} 
    \includegraphics[width=\linewidth]{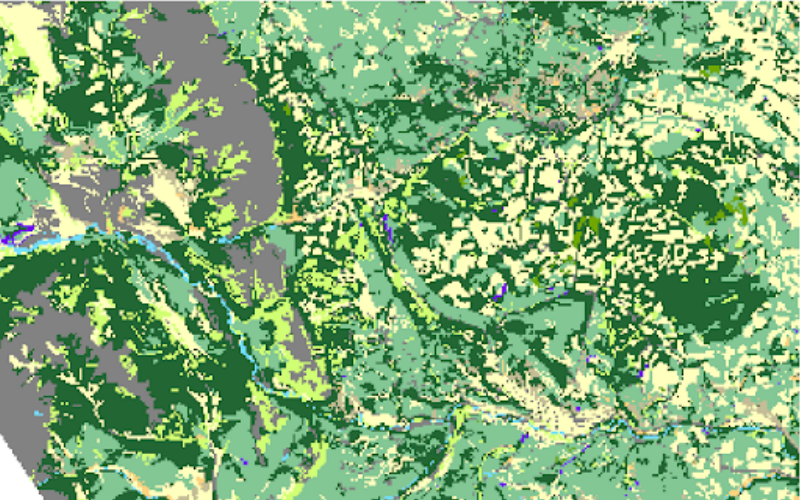}
    \caption{Original landscape.}
    \label{fig:OrigForest}
    \end{minipage}
    \hspace{1.5cm}
    \begin{minipage}[b]{0.34\textwidth} 
    \includegraphics[width=\linewidth]{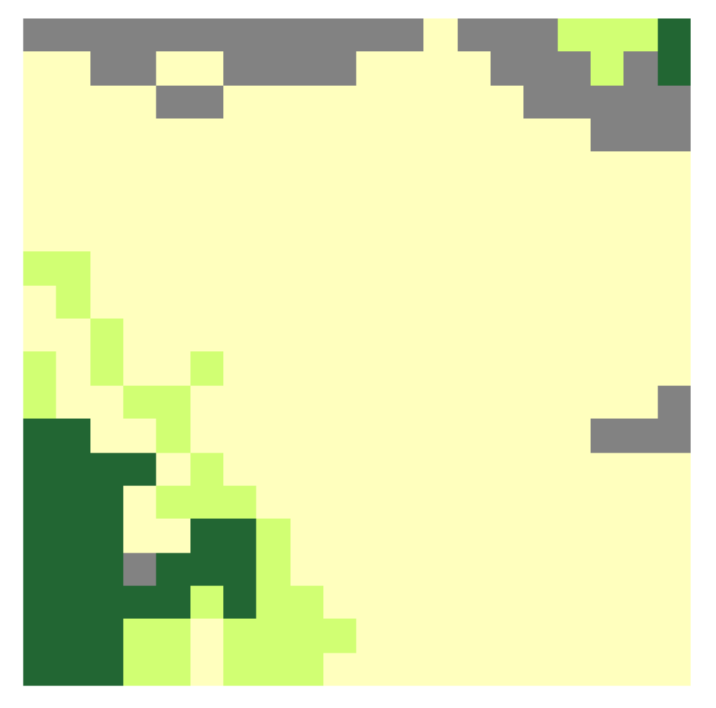}
    \caption{\texttt{Sub20} landscape.}
    \label{fig:Sub20}
  \end{minipage}
\end{figure}

\begin{figure}[h!]
\centering
    \hspace{-1.5cm}
    \begin{minipage}[b]{0.34\textwidth}
    \includegraphics[width=\linewidth]{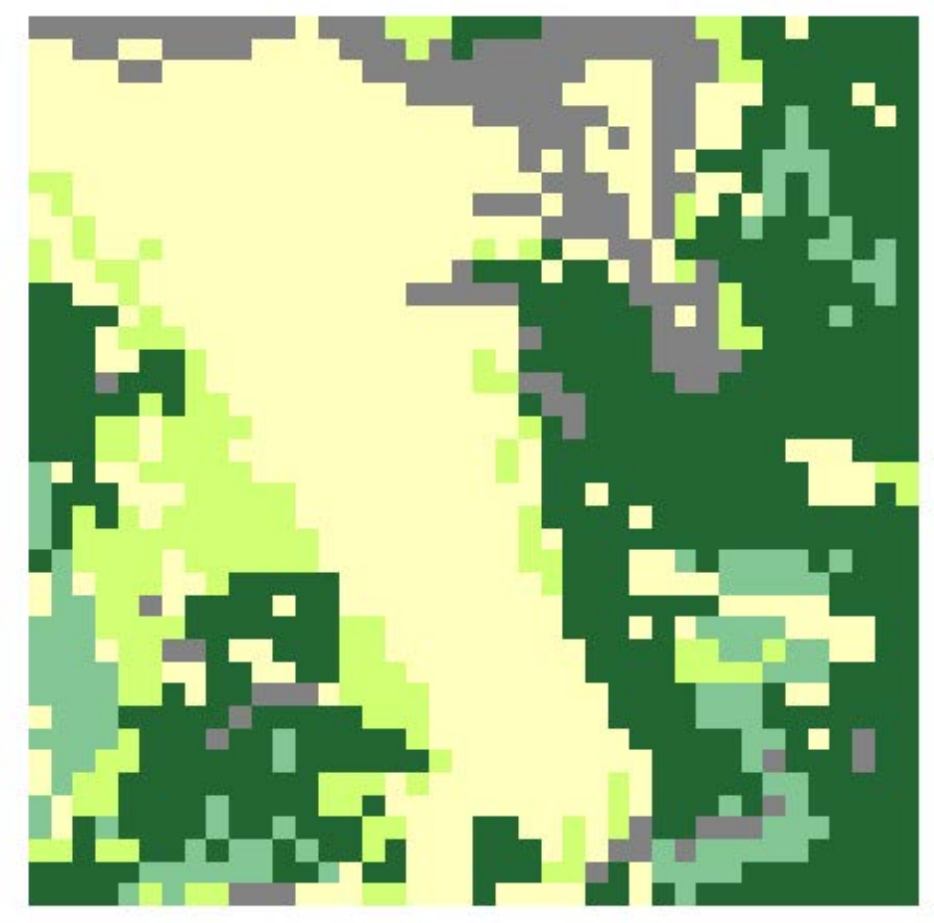}
    \caption{\texttt{Sub40} landscape.}
    \label{fig:Sub40}
  \end{minipage}
  \hspace{2.5cm}
  \begin{minipage}[b]{0.24\textwidth}
    \includegraphics[width=\linewidth]{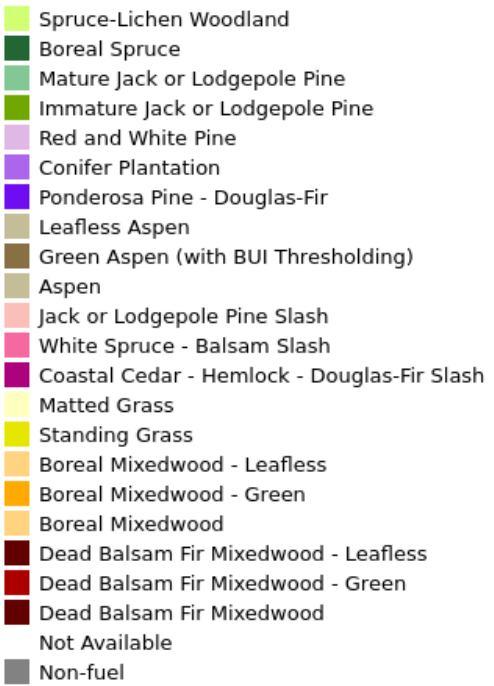}
    \caption{Fuel types.}
    \label{fig:fuel-types}
  \end{minipage}
\end{figure}

We employed two real landscapes in our experiments: \texttt{Sub20} and  \texttt{Sub40}, both situated in the Alberta region of Canada. The original instance is shown in Fig.\:\ref{fig:OrigForest}. The \texttt{Sub20} landscape, a 400-hectare forest patch depicted in Fig.\:\ref{fig:Sub20}, was chosen for assessing computational performance and observing how the objective function evolves with changes in model parameters. To evaluate the model's computational efficacy over larger forest areas, we extended the study to include \texttt{Sub40} (Fig.\:\ref{fig:Sub40}) under specific parameter configurations. Each landscape is divided into $100 \times 100$ m$^2$ cells, each containing a specific fuel (Fig.\:\ref{fig:fuel-types}), detailed information about these landscapes can be found at \url{https://github.com/fire2a/C2FFBP}. Simulations of multiple wildfires were conducted using Cell2Fire, requiring fire-weather scenarios specific to the study area \citep{parisien2005mapping, pais2021cell2fire}. These scenarios include crucial factors such as temperature, relative humidity, wind speed, wind direction, and fire weather indices-essential inputs for the Canadian Fire Behavior Prediction (FBP) System \citep{hirsch1996canadian}.

Going into more detail, in \texttt{Sub20} ignition points were delimited to a circumference of a radio of 4 cells at the center of the forest. Regarding weather scenarios, these were constructed using real data comprising the zone of interest. Without any fire treatment, on average around 18\% of the total number of cells are burned. 

\begin{figure}
    \centering
    \begin{minipage}[b]{0.38\textwidth}
        \centering
        \includegraphics[width=\textwidth]{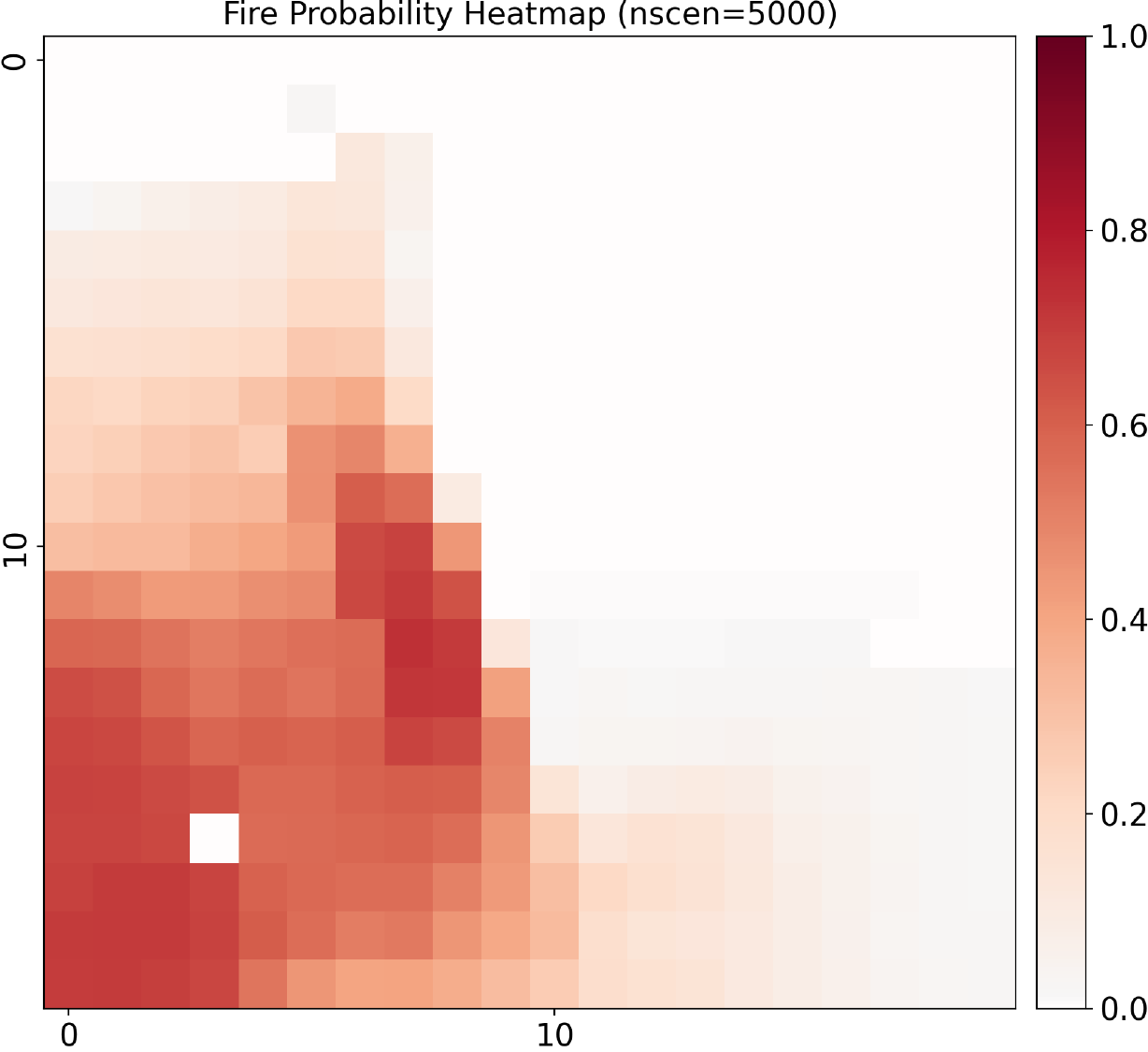}
        \subcaption[]{\texttt{Sub20}}
        \label{fig:fire-pre-a}
    \end{minipage}
    \hspace{1cm}
    \begin{minipage}[b]{0.38\textwidth}
    \centering
        \includegraphics[width=\textwidth]{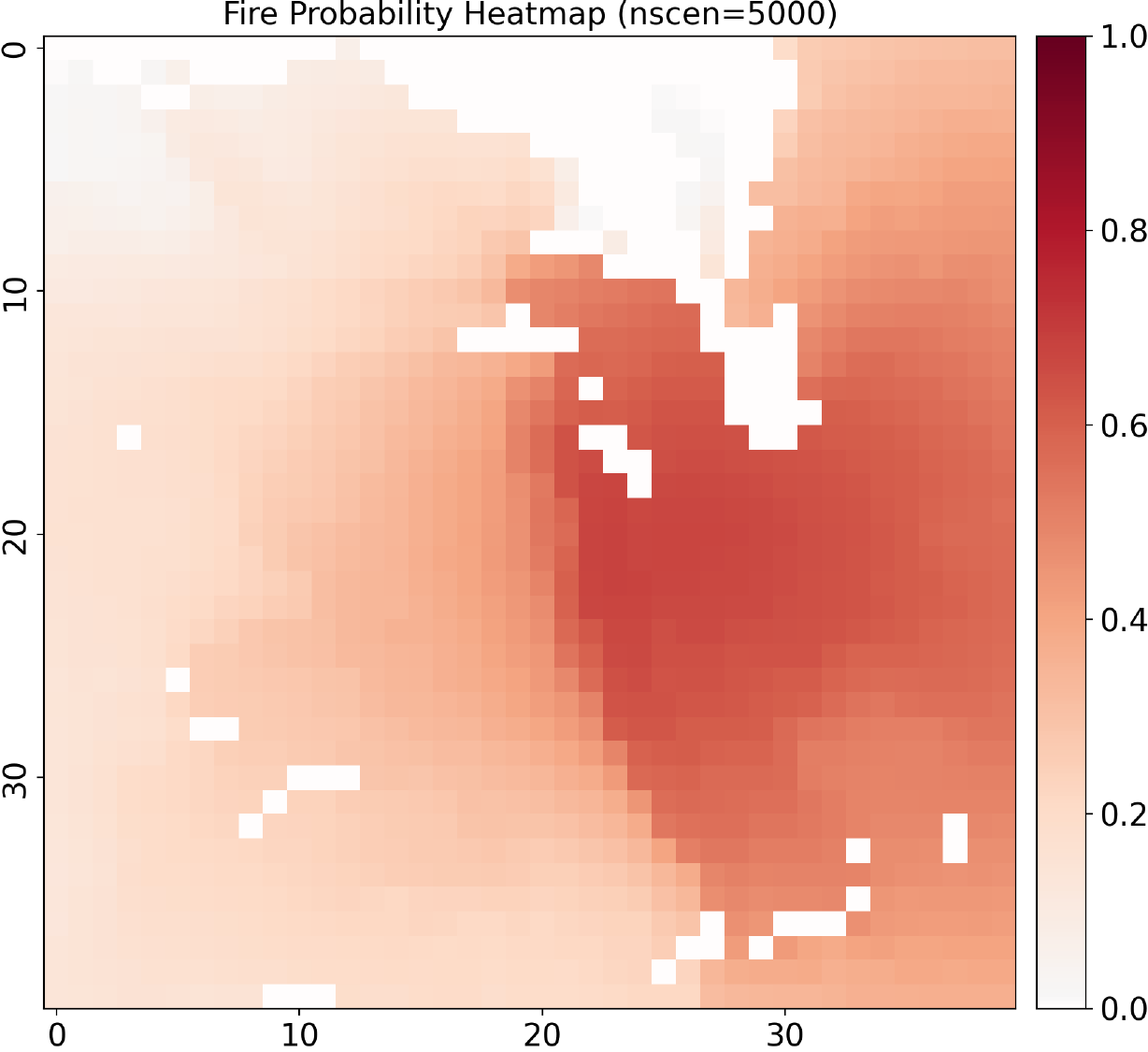}
        \subcaption[]{\texttt{Sub40}}
        \label{fig:fire-pre-b}
    \end{minipage}   
    \caption{Fire spreading patterns for both instances.}
    \label{fig:fire-pre}
\end{figure}

Similarly for \texttt{Sub40}, ignitions were delimited to a circumference of radio 9 cells at the center of the forest. Without any intervention, close to 31\% of the total number of cells are burned. The general spreading pattern of both instances is reflected in Fig.\:\ref{fig:fire-pre}.

This larger forests were then shrank to generate smaller forest that exhibit similar behavior of size $10\times10$, in order to search efficiently in the hyperparameter space. An interpolation method designed for image shrinking was used, called Nearest-neighbor Interpolation, where the value of each pixel in the resulting image is equal to the one the algorithm determines to be the closest.

As was mentioned in section \ref{SS:Algorithms}, learning from demonstration was used as a way of guiding the agent's behavior towards better sections of the action space faster. In order to generate demonstrations, a baseline algorithm was constructed that makes use of a widely used landscape connectivity metric called Downstream Protection Value (DPV), presented in \cite{pais2021downstream}. This measure considers a forest as an undirected graph $G=(V,E)$, in which each node corresponds to a cell and edges to contiguous cells. Then a number of fire spreading simulations are run, each generating a directed sub graph $G_{d}=(V_{d},E_{d})$ over $G$, where $V_{d}$ contains all nodes that were burned and $E_{d}$ all edges that transmitted fire. With this sub graph, the minimum spanning tree is calculated for each node as root $T_{d}(j)=(V_{d}(j), E_{d}(j))$ and the DPV for cell $i$ is defined as:

\begin{equation} \label{Eq:DPV}
    DPV(i) = \frac{1}{|D|} \sum_{d \in D}\sum_{j \in V_{d}(i)} v(j)
\end{equation}

where $v(j)$ is some value at risk for a node $j$ and $D$ is a set of simulations. To this date, the DPV is among the top metrics to solve the FBP problem. Finally, the algorithm used as demonstrator, called baseline, was constructed and is presented in Algorithm.\:\ref{alg:dpv}.

\begin{algorithm}[h]
\caption{Baseline}\label{alg:dpv}
\begin{algorithmic}
\For{episode $\in$ episodes}
    \State $t' \gets 0$
    \For{$t \leq T$}
        \State Run $D$ simulations over $s_{t}$ using Cell2Fire
        \State Calculate $DPV$ for each $i \in V$
        \State $a_{t} \gets argmax_{i \in V} DPV(i)$
        \State Execute $a_{t}$ and observe $(r_{t}, s_{t+1})$
        \State Store transition $(s_{t}, a_{t}, r_{t}, s_{t+1})$     
    \EndFor
\EndFor

\end{algorithmic}
\end{algorithm}

 Hyperparameter tuning was done performing a grid search using Deep Q-Learning with 1000 demonstrations over a shrinked version of the first forest of size $10\times10$, which resulted in the values shown in Table \ref{parameter-table}. This process was done for small-net and big-net. For the pre-trained models those parameters found for big-net were used since it is the most similar architecture in terms of number of parameters and replicating the experiments was highly time consuming.

\section{Results and Discussion}
\label{S:Results}

\subsection{\texttt{Sub20}}
\label{SS:R-20}
In the analysis of the \texttt{Sub20} landscape, the outcomes are highly satisfactory. This study systematically evaluates various algorithms across different network architectures, revealing that all tested algorithms not only converge but also outperform the baseline model detailed in \ref{SS:Experimental-Setup}. Significantly, these algorithms yield rewards that greatly exceed those of a random strategy, as depicted by the ``random" curve in our comparative analysis.

\begin{figure*}[h!]
    \centering
  \begin{minipage}[b]{0.45\textwidth}
    \includegraphics[width=\linewidth]{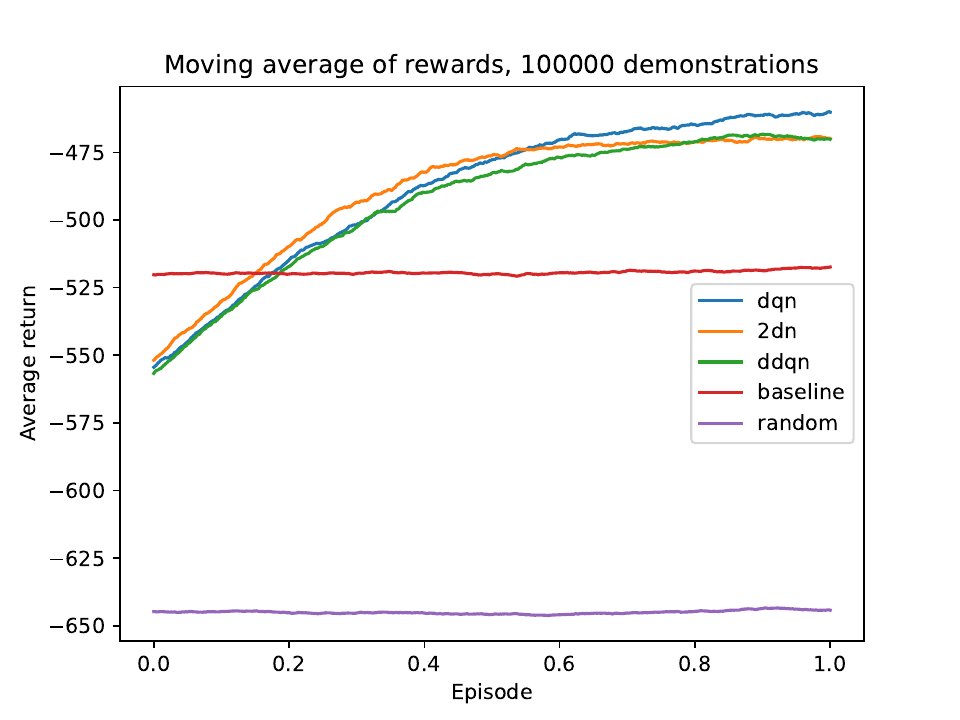}
    \subcaption[]{small-net}
    \label{fig:curves_hetero1_small}
    \end{minipage}
  \begin{minipage}[b]{0.45\textwidth}
    \includegraphics[width=\linewidth]{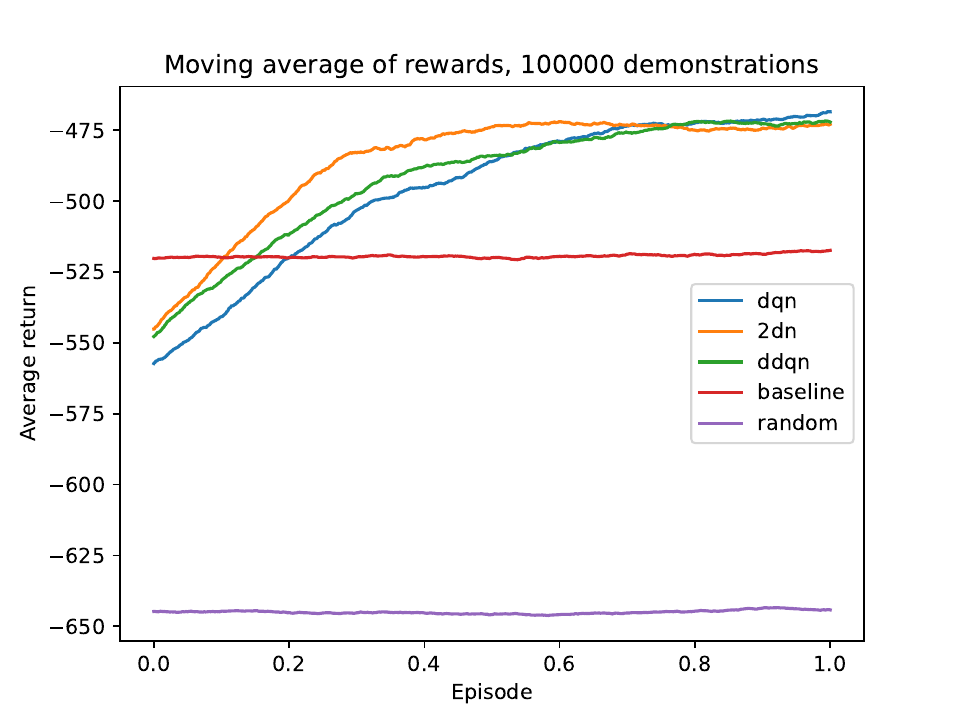}
    \subcaption[]{big-net}
    \label{fig:curves_hetero1_big}
    \end{minipage}
  \begin{minipage}[b]{0.45\textwidth}
    \includegraphics[width=\linewidth]{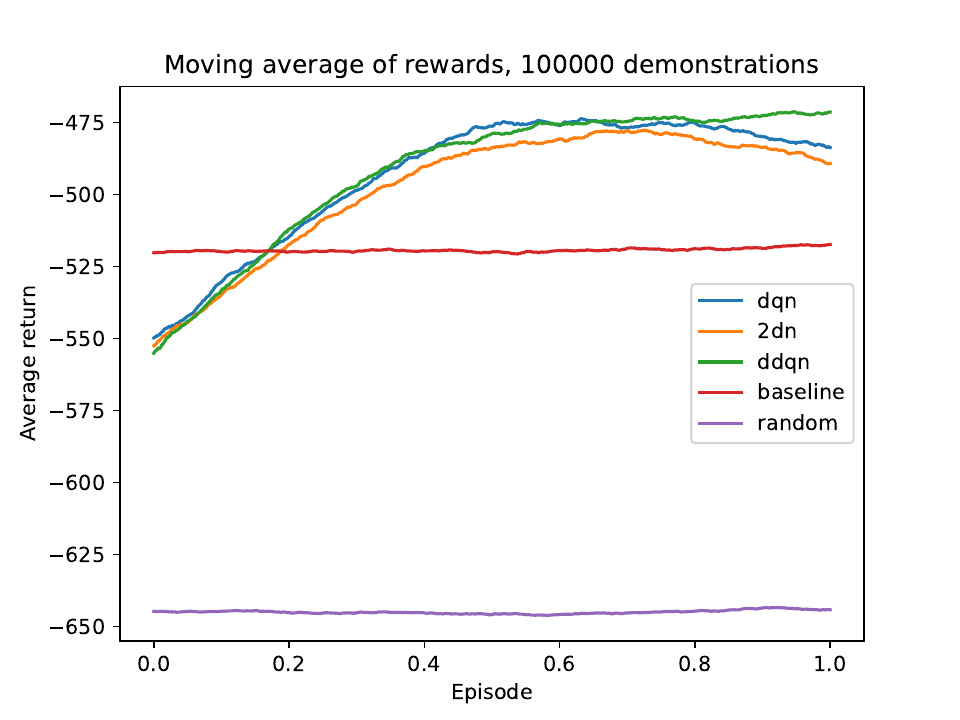}
    \subcaption[]{small-mobile}
    \label{fig:curves_hetero1_small-mobile}
    \end{minipage}
  \begin{minipage}[b]{0.45\textwidth}
    \includegraphics[width=\linewidth]{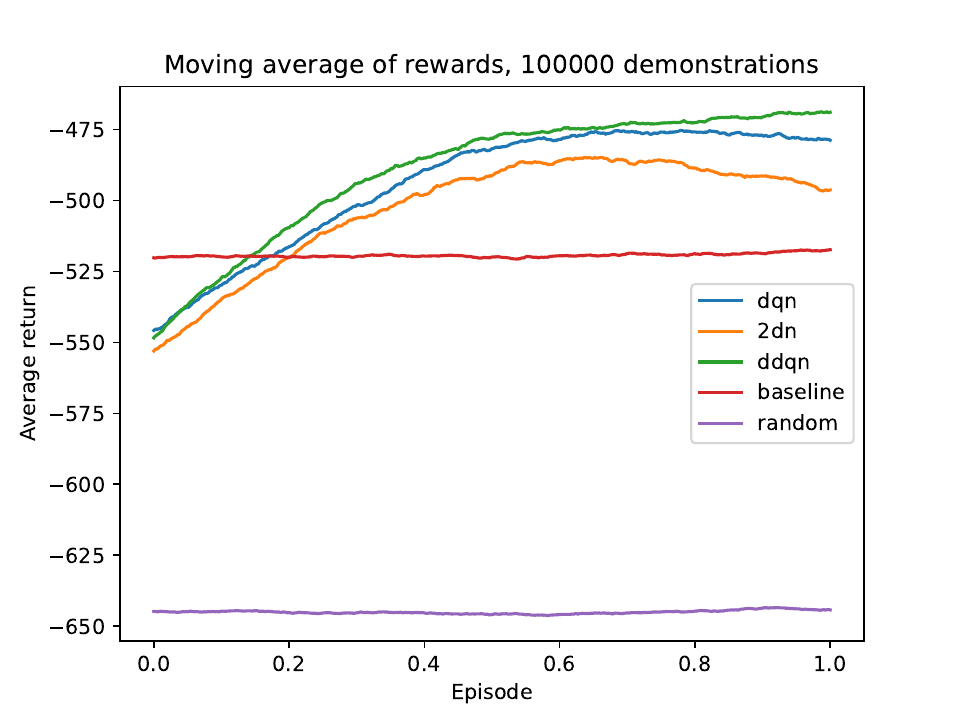}
    \subcaption[]{big-mobile}
    \label{fig:curves_hetero1_big-mobile}
    \end{minipage}
  \begin{minipage}[b]{0.45\textwidth}
    \includegraphics[width=\linewidth]{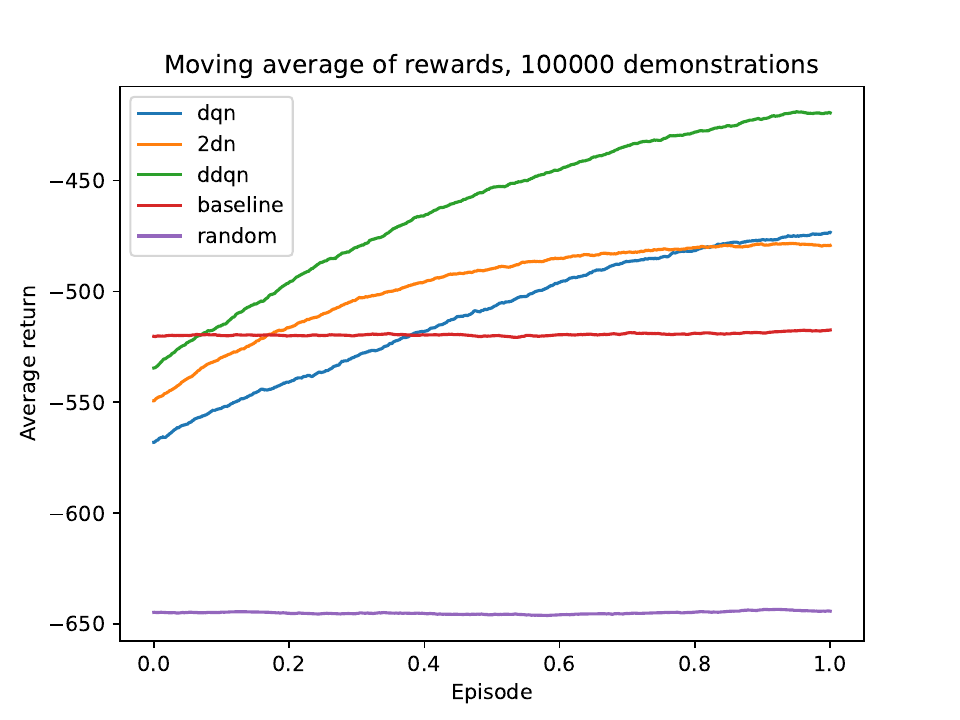}
    \subcaption[]{efficient-net}
    \label{fig:curves_hetero1_efficient-net}
    \end{minipage}
    \caption{Learning curves for the \texttt{Sub20} instance.}
    \label{fig:curves_hetero1}
\end{figure*}

A detailed qualitative examination of the reward curves in Fig.\:\ref{fig:curves_hetero1}
 suggests that the three primary algorithms under consideration perform comparably, with no single algorithm consistently outshining the others across all scenarios. This trend holds true across diverse architectures, except for efficient-net. Efficient-net uniquely demonstrates a marginal superiority in reward acquisition.

\begin{figure}
    \centering
    \begin{minipage}[b]{0.48\textwidth}
        \includegraphics[width=\textwidth]{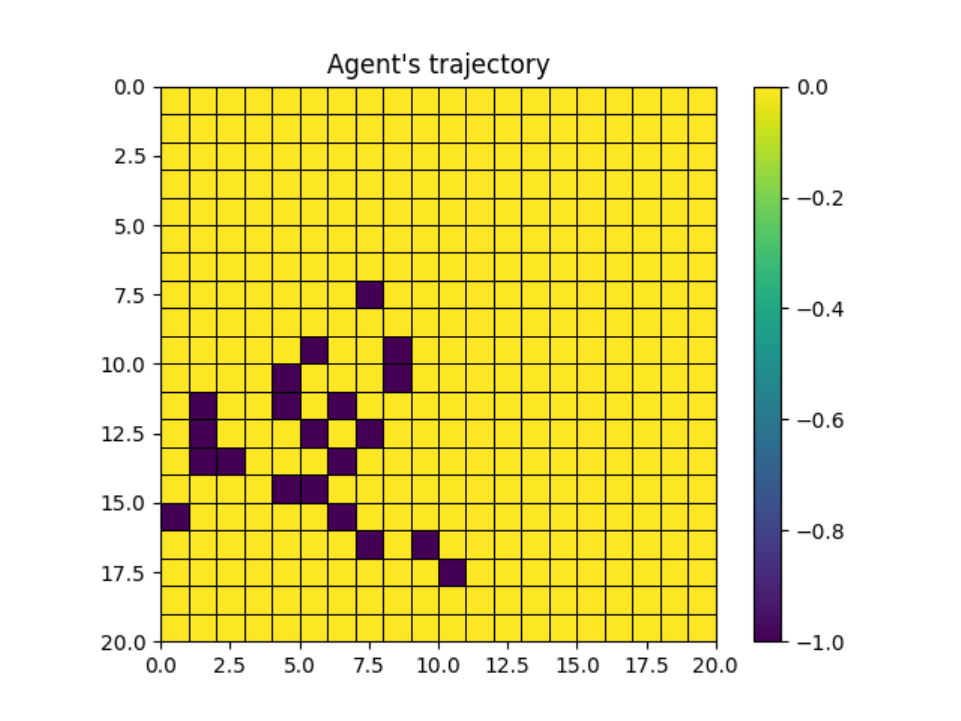}
        \vspace{-0.8cm}
        \subcaption[]{Agent's trajectory}
        \label{fig:solution_hetero_1}
        
    \end{minipage}
    \begin{minipage}[b]{0.48\textwidth}
    \centering
        \includegraphics[width=0.68\textwidth]{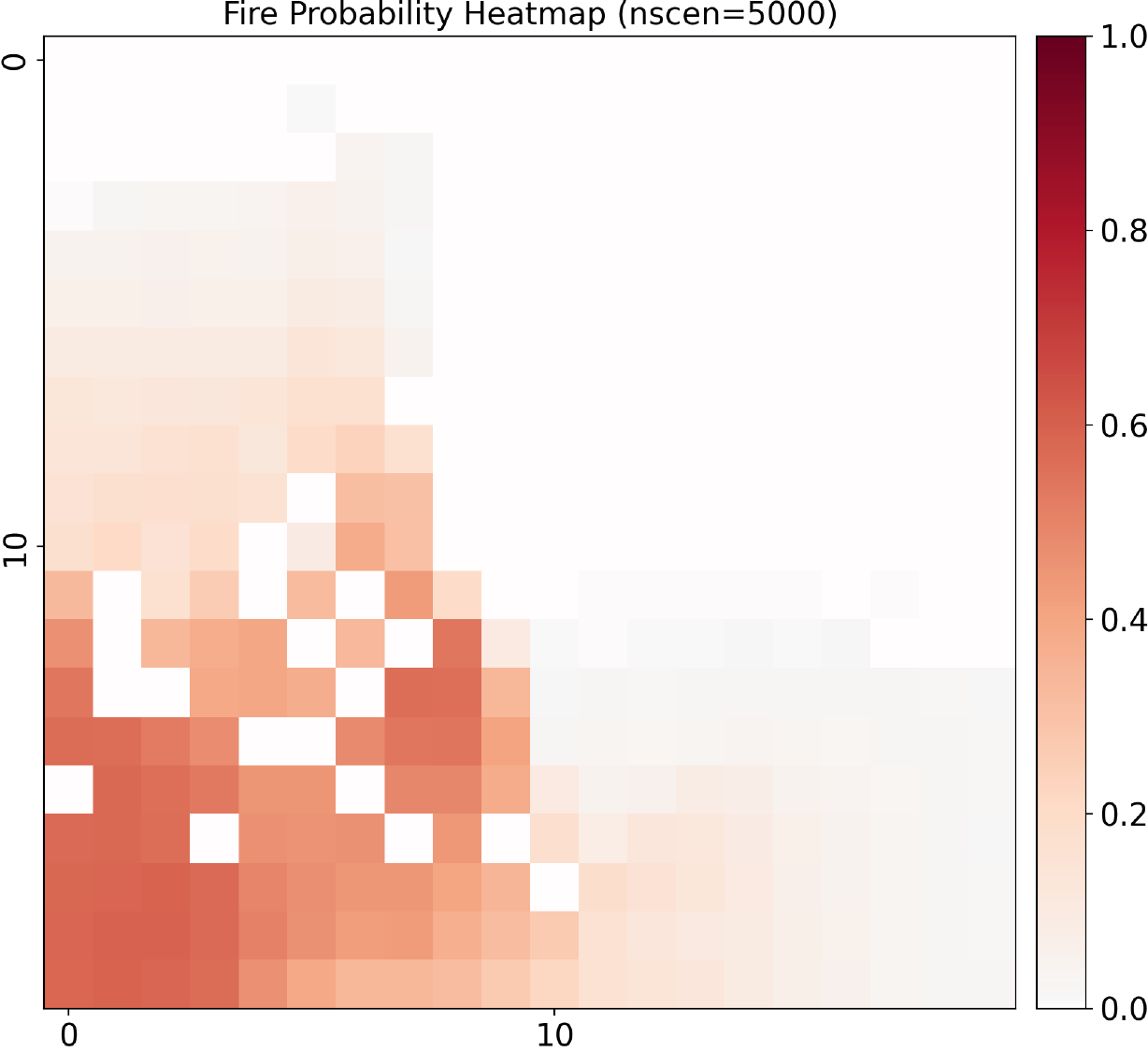}
        \subcaption[]{Burn Probability Map}
        \label{fig:spread_hetero_1}
    \end{minipage}   
    \caption{Best solution obtained and the corresponding fire spreading behavior for the \texttt{Sub20} forest.}
    \label{fig:solution-hetero-1}
\end{figure}

Focusing on the aspect of fire spread mitigation, a qualitative comparison between the burn probability maps of the untreated landscape (Fig.\:\ref{fig:fire-pre}) and the landscape post-treatment (Fig.\:\ref{fig:solution-hetero-1}) reveals a substantial reduction in fire spread. This is further quantified by examining the percentage of burned cells: under a random solution, 16.1\% of the forest succumbs to fire, while the baseline algorithm limits this to 12.9\%. In contrast, the solutions derived from DRL algorithms exhibit a narrower range of 11.31\% to 12.86\% in terms of burned forest area. These percentages correspond to the outcomes of the Deep Q-Learning (DQN) algorithm using the small-net architecture and the Dueling Double Deep Q-Learning (DDQN) algorithm implemented with efficient-net, respectively.
\begin{table*}[h] 
      \begin{tabular*}{\linewidth}{@{\extracolsep{\fill}}l|lllll}
    
               & small-net & big-net  & small-mobile & big-mobile & efficient-net \\ \hline
    DQN         & 11.31   & 11.44 & 12.47       & 11.78    & 11.44\\ \hline
    2DQN       & 11.6     & 11.63 & 12.51       & 12.11    & 11.6\\ \hline
    DDQN       & 11.6     & 11.6 & 11.6       & 11.85    & 12.86\\ \hline
    baseline     & \multicolumn{5}{l}{12.9}  \\ \hline
    random     & \multicolumn{5}{l}{16.1}  \\ 
    \end{tabular*}
    \caption{\label{demo-table} Porcentage of landscape burned after treatment in \texttt{Sub20} instance.}
    \label{table-burn-1}
\end{table*}

\subsection{\texttt{Sub40}}
\label{SS:R-40}
\begin{figure*}[h!]
    \centering
  \begin{minipage}[b]{0.45\textwidth}
    \includegraphics[width=\linewidth]{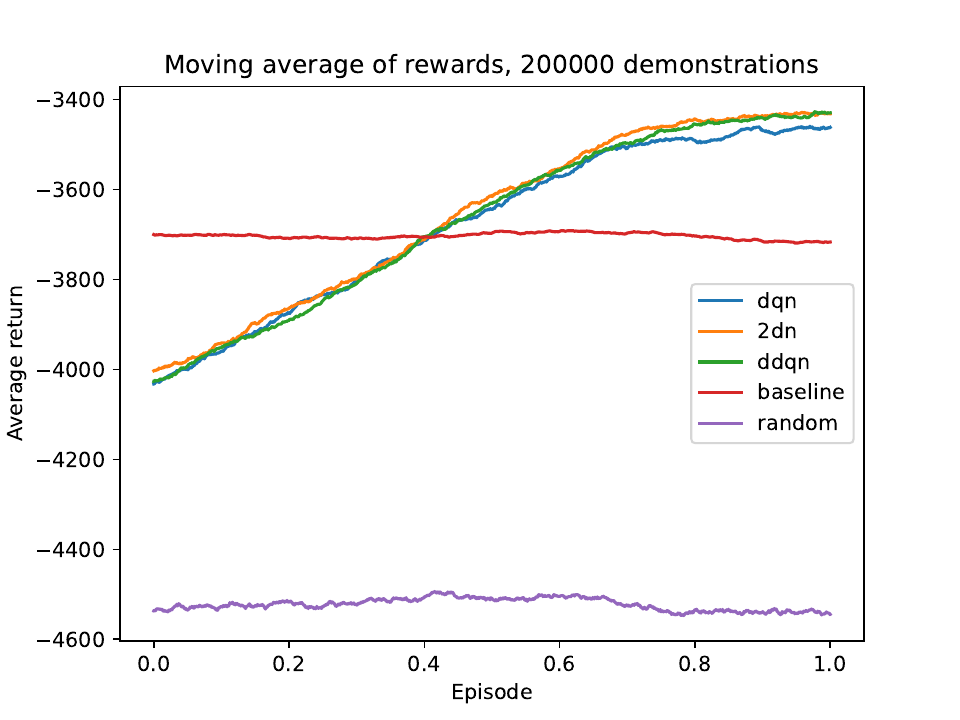}
    \subcaption[]{small-net}
    \label{fig:curves_hetero_2_small}
    \end{minipage}
  \begin{minipage}[b]{0.45\textwidth}
    \includegraphics[width=\linewidth]{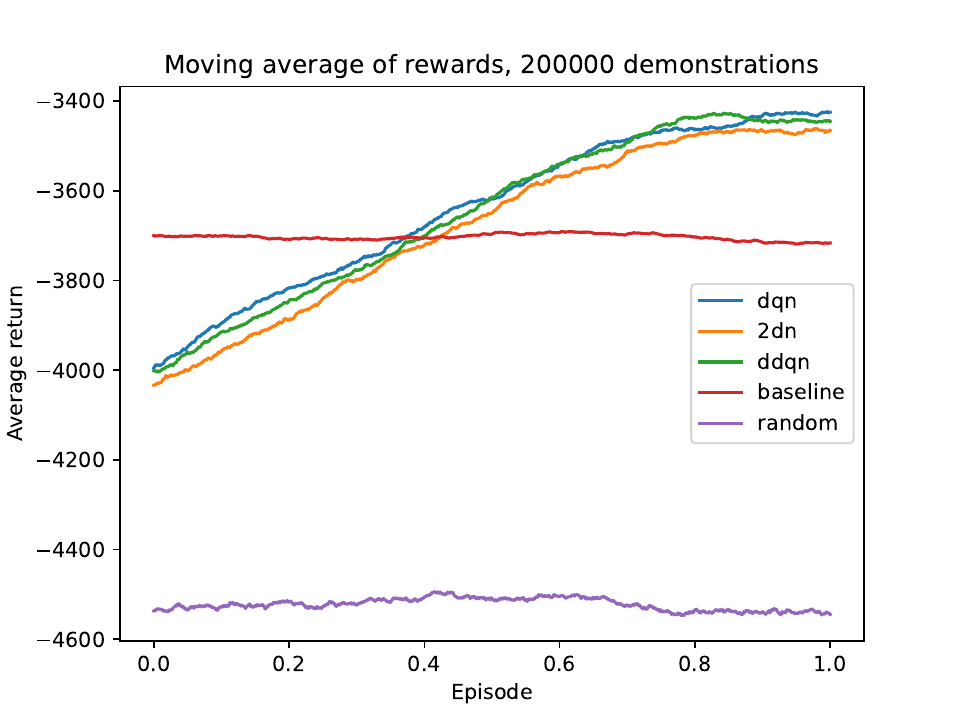}
    \subcaption[]{big-net}
    \label{fig:curves_hetero_2_big}
    \end{minipage}
  \begin{minipage}[b]{0.45\textwidth}
    \includegraphics[width=\linewidth]{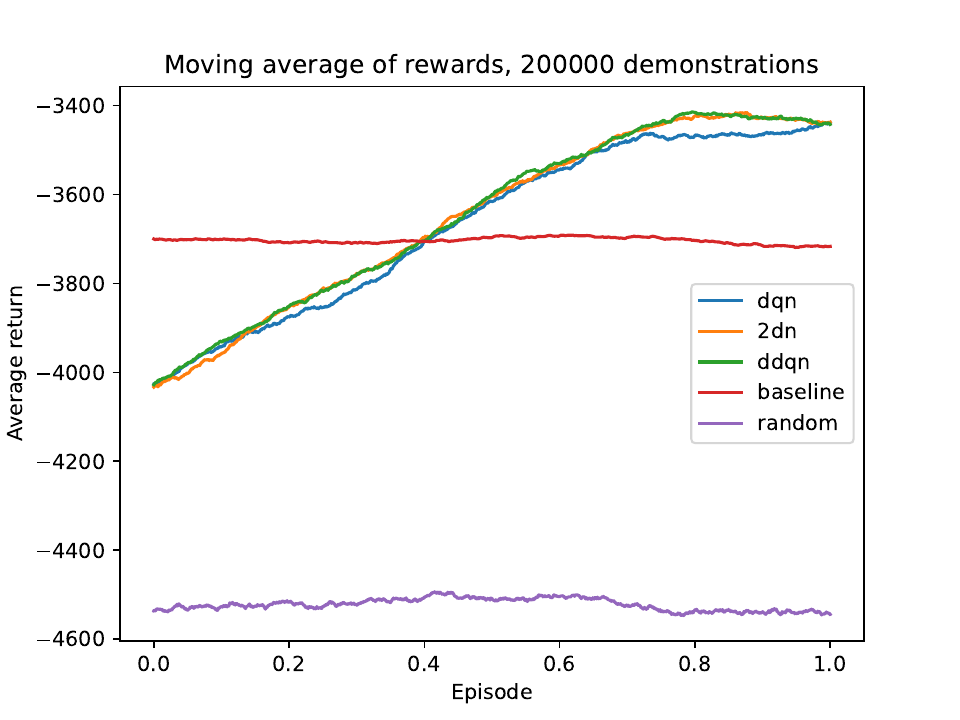}
    \subcaption[]{small-mobile}
    \label{fig:curves_hetero_2_smal-mobile}
    \end{minipage}
  \begin{minipage}[b]{0.45\textwidth}
    \includegraphics[width=\linewidth]{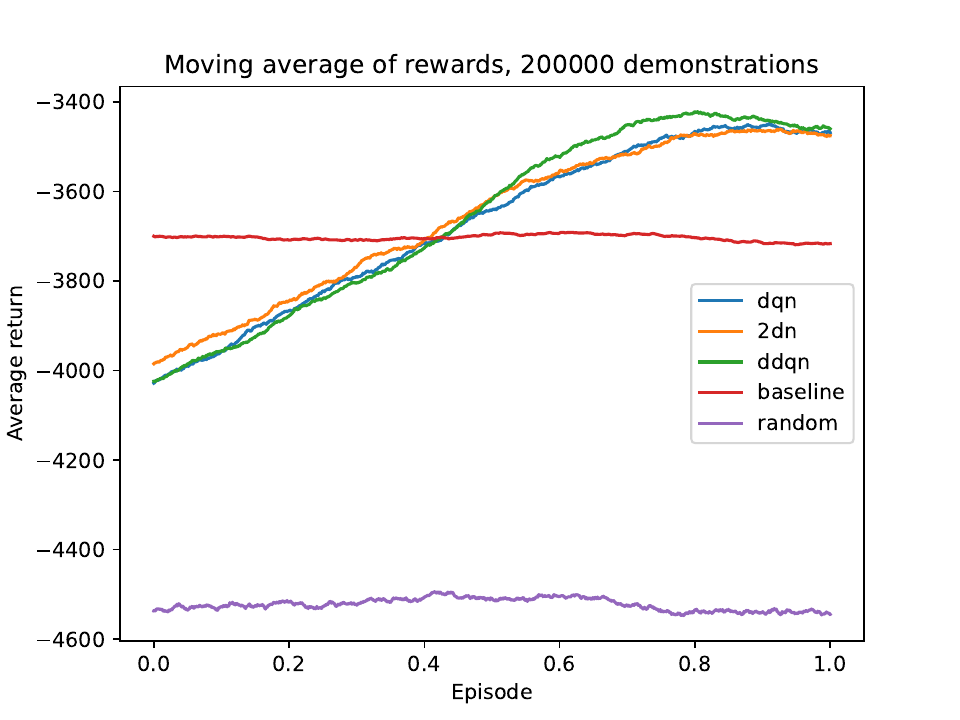}
    \subcaption[]{big-mobile}
    \label{fig:curves_hetero_2_big-mobile}
    \end{minipage}
  \begin{minipage}[b]{0.45\textwidth}
    \includegraphics[width=\linewidth]{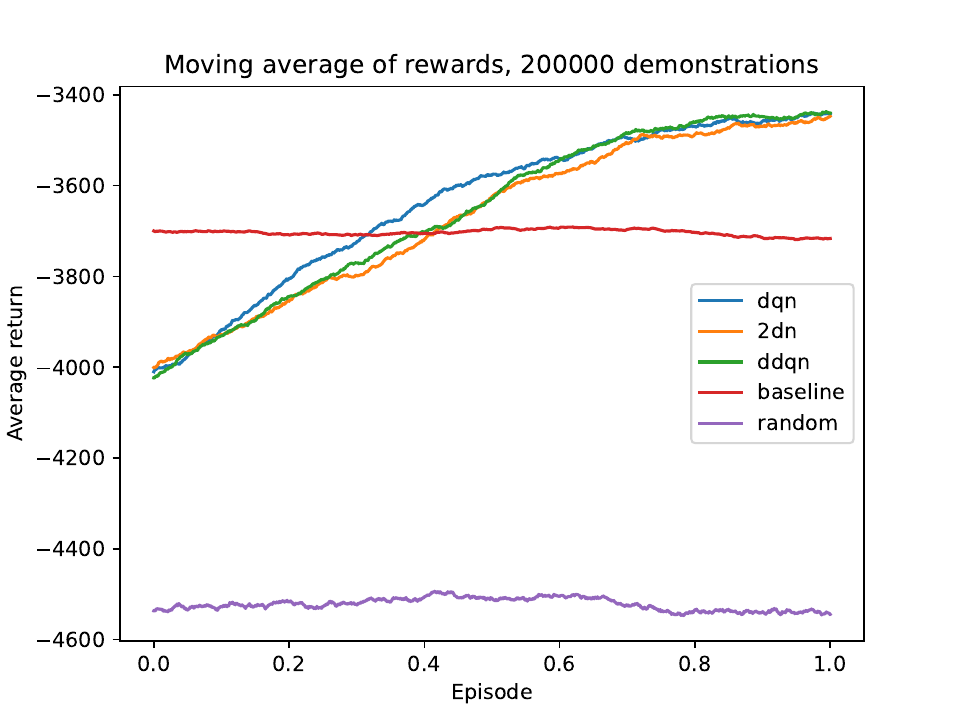}
    \subcaption[]{efficient-net}
    \label{fig:curves_hetero_2_efficient-net}
    \end{minipage}
    \caption{Learning curves for the \texttt{Sub40} instance.}
    \label{fig:curves_hetero_2}
\end{figure*}

In the context of the \texttt{Sub40} forest simulation, the results are notably positive. A comparative analysis across five distinct network architectures reveals that all three employed algorithms not only achieve convergence but also significantly outperform the baseline algorithm. This superiority is evident in the reward metrics in Fig.\:\ref{fig:curves_hetero_2}, where the DRL algorithms attain reward levels close to -3.400, markedly better than the -3.700 achieved by the demonstrator algorithm. Moreover, these algorithms also surpass a random algorithm, which records a lower reward level of approximately -4.500. Interestingly, the performance of the three DRL algorithms is relatively comparable, with no single algorithm demonstrating a substantial edge over the others.

\begin{figure*}
    \centering
    \begin{minipage}[b]{0.48\textwidth}
    \centering
        \includegraphics[width=\textwidth]{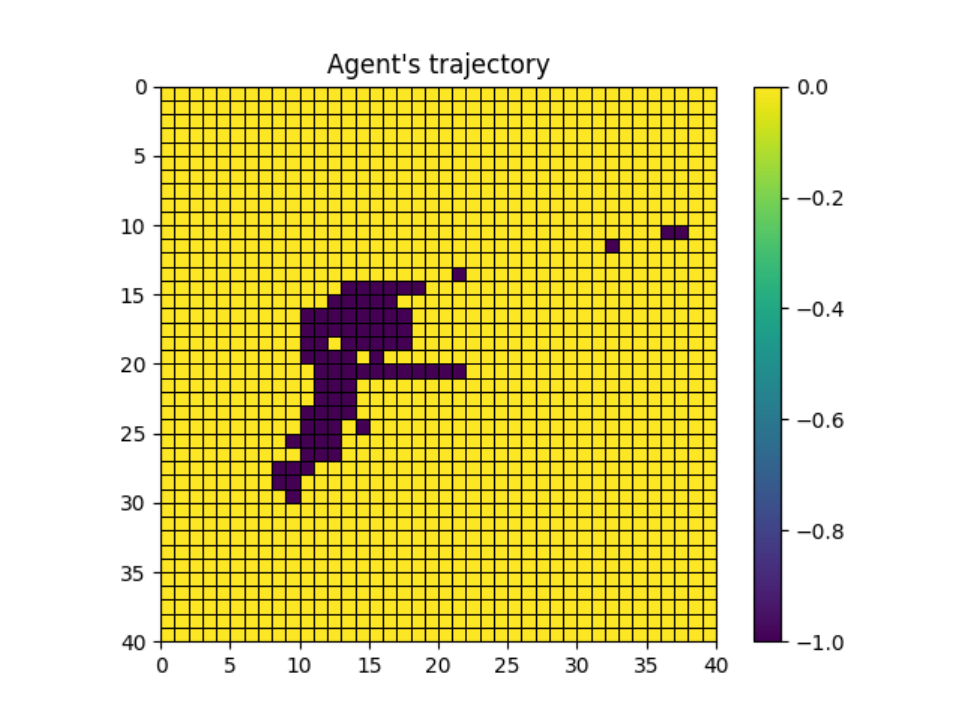}
        \vspace{-0.8cm}
        \subcaption[]{Agent's trajectory}
        \label{fig:solution_hetero_2}
    \end{minipage}
    \begin{minipage}[b]{0.48\textwidth}
    \centering
        \includegraphics[width=0.68\textwidth]{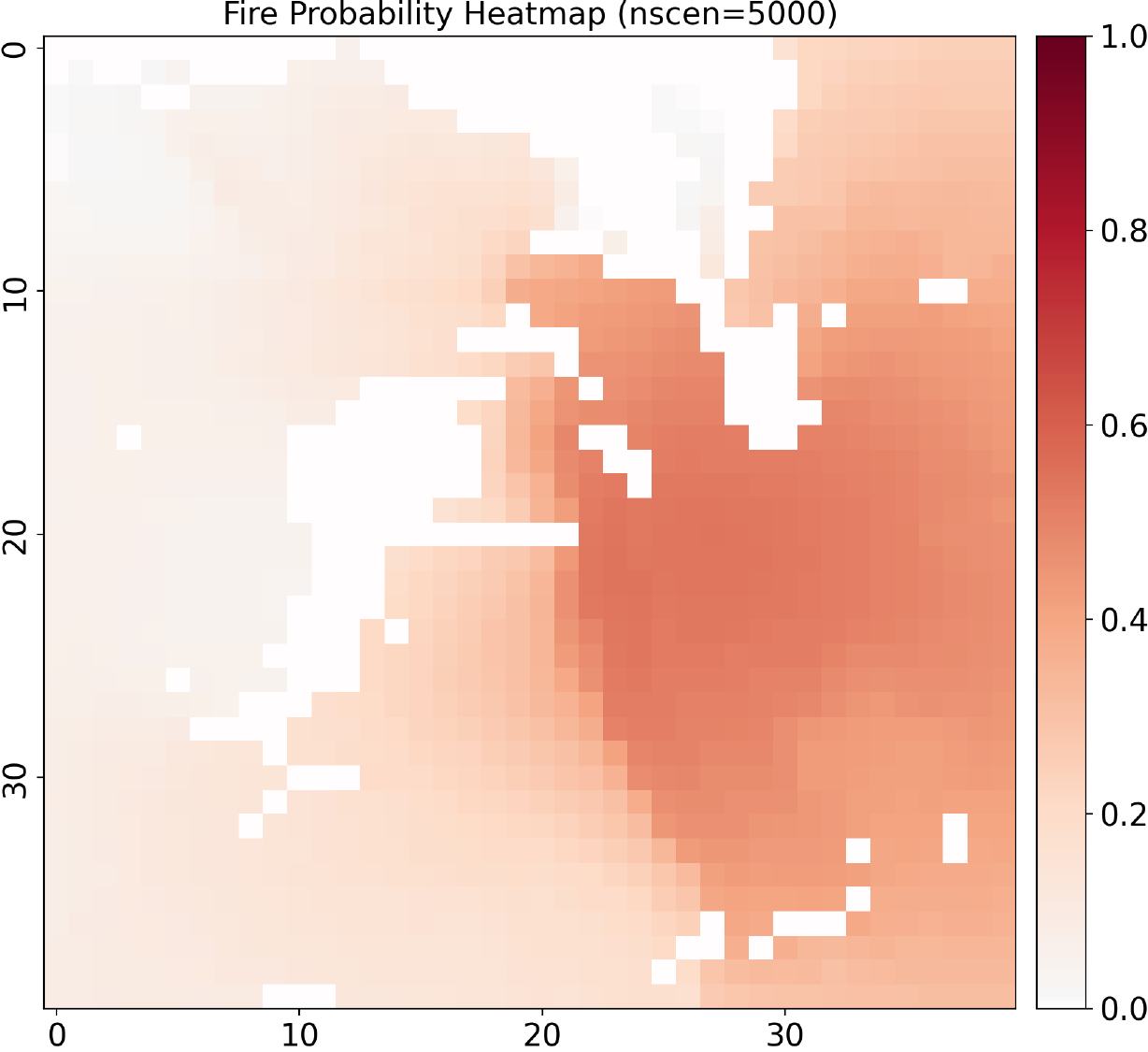}
        \subcaption[]{Burn Probability Map}
        \label{fig:spread_hetero_2}
        
    \end{minipage}   
    \caption{Best solution obtained and the corresponding fire spreading behavior for the \texttt{Sub40} forest.}
    \label{fig:solution-hetero2}
\end{figure*}

Regarding the aspect of fire spread control, the DRL algorithms demonstrate a substantial reduction in the number of burned cells within the simulation. This efficacy is visually represented in Fig.\:\ref{fig:solution-hetero2}'s burn probability map, which exhibits lower values compared to the untreated landscape depicted in Fig.\:\ref{fig:fire-pre}. Notably, Table \ref{table-burn-2}. presents quantitative data where the percentage of burned cells is significantly lower for landscapes managed by the DRL algorithms compared to those managed by the baseline and a random algorithm. Specifically, while the baseline and random algorithms result in 23.25\% and 28.36\% of the total area being burned, respectively, the DRL algorithms, particularly the Deep Q-Learning (DQN) with efficient-net, limit this to a range of 21.55\% to 21.78\%. These findings underscore the effectiveness of DRL algorithms in mitigating fire spread in forest simulations.

\begin{table*}[h]
    \begin{tabular*}{\linewidth}{@{\extracolsep{\fill}}l|lllll}
    
               & small-net & big-net  & small-mobile & big-mobile & efficient-net \\ \hline
    DQN        & 21.73   & 21.64 & 21.78       & 21.78    & 21.55\\ \hline
    2DQN       & 21.78     & 21.78 & 21.78       & 21.78    & 21.78\\ \hline
    DDQN       & 21.78     & 21.78 & 21.78       & 21.78    & 21.74\\ \hline
    baseline     & \multicolumn{5}{l}{23.25}  \\ \hline
    random     & \multicolumn{5}{l}{28.36}  \\ 
    \end{tabular*}
    \caption{\label{demo-table} Porcentage of landscape burned after treatment in \texttt{Sub40} instance.}
    \label{table-burn-2}
\end{table*}

Another point to emphasize is that the results obtained in these applications do not show any consistent improvement through the introduction of network complexity. Specifically, small-net performs just as well as big-net, suggesting that the task at hand may not require increased model complexity. This notion is further supported by the observation that using transfer learning yields similar results to not using it. This could indicate that 1) the proposed architectures (small-net and big-net) are sufficiently deep and capable of extracting the necessary features for the task, or 2) it might suggest that too many layers of the pre-trained networks were 'frozen', preventing them from improving the results or 3) that the tasks for which the pre-trained models were trained for are not similar enough to the actual task at hand and therefore their usage is of no real convenience.

Upon initial examination, the results suggest a greater improvement for \texttt{Sub40} compared to \texttt{Sub20}. However, a closer analysis reveals a nuanced picture: the reduction in burned cells is 30\% for \texttt{Sub40} and slightly higher at 37\% for \texttt{Sub20}. This observation indicates that the performance across both cases is similarly effective, despite the problem's complexity increasing exponentially with size. A plausible explanation for this counter intuitive finding could be that the more complex spreading patterns provide richer information inputs. These inputs potentially enable the agent to derive more sophisticated insights, offsetting the challenges posed by increased complexity.

%\iffalse
\subsection{Understanding the agent}
\label{SS:Understanding}
As previously discussed in \ref{SS:Explainability}, the models employed in this study inherently lack self-explanatory capabilities, this requiring the integration of explainability techniques to elucidate their predictive outcomes. This is particularly crucial in our application where understanding whether the model accounts for the impact of existing firebreaks is vital, a factor often overlooked in conventional methods. To address this, we generated attention maps using the methodology outlined in \ref{SS:Explainability}.

\begin{figure}[pos=h]
    \centering
    \begin{minipage}[b]{0.45\textwidth}
    \centering
    \includegraphics[width=0.9\textwidth]{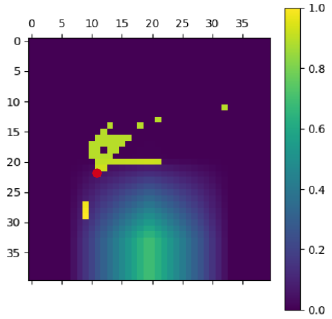}
    \subcaption[]{}
    \label{fig:interp_1}
    \end{minipage}
    \begin{minipage}[b]{0.45\textwidth}
    \centering
    \includegraphics[width=0.9\textwidth]{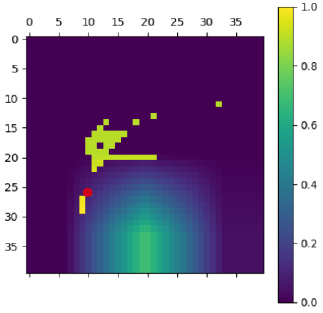}
    \subcaption[]{}
    \label{fig:interp_2}
    \end{minipage}
    \begin{minipage}[b]{0.45\textwidth}
    \centering
    \includegraphics[width=0.9\textwidth]{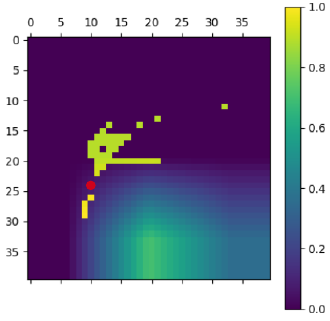}
    \subcaption[]{}
    \label{fig:interp_3}
    \end{minipage}
    \begin{minipage}[b]{0.45\textwidth}
    \centering
    \includegraphics[width=0.9\textwidth]{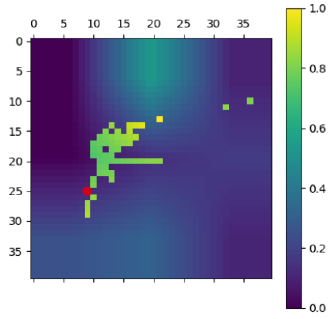}
    \subcaption[]{}
    \label{fig:interp_4}
    \end{minipage}
    \caption{Decision making process while solving \texttt{Sub40}.}
    \label{fig:decision-grads}
\end{figure}

These maps, specifically from Double Deep Q-Learning models using efficient-net (Fig.\:\ref{fig:decision-grads}), were chosen to demonstrate the model's consideration of existing firebreaks, though they should not be viewed as indicative of overall results. The attention maps overlay the current forest state, with the red dot marking the proposed firebreak location. These maps highlight the model's focus areas when estimating the state-action function $q(s,a,\theta)$, with higher values indicating greater attention.

Figures \:\ref{fig:interp_1}, \:\ref{fig:interp_2} and \:\ref{fig:interp_3} illustrate consecutive decisions by the agent, reflecting a strategy to enclose a specific forest area. The model appears to prioritize cells safeguarded by the cumulative effect of the new and existing firebreaks, particularly in the lower-central region, suggesting an awareness of the synergistic effect with existing firebreaks. This observation tentatively supports the hypothesis that the model comprehends the complex interactions inherent in firebreak allocation.

However, caution is advised in interpreting these results due to their subjective nature. Fig.\:\ref{fig:interp_4} exemplifies this, showing the model's focus on seemingly unrelated areas for firebreak placement. This discrepancy underscores the need for cautious interpretation and further investigation into the model's decision-making process.
%\FloatBarrier
%\fi

\section{Conclusions and Future Work}
\label{S:Conclusions}
The primary motivation of this work was to assess the feasibility of using DRL to solve a challenging problem in forest engineering. This goal was achieved, under certain conditions. Firstly, considering the algorithms used, the incorporation of demonstrations is crucial for obtaining good results as shown in Fig.\:\ref{fig:no-demonstrations}. Without them, neither convergence nor good performance is achieved. This could be due to various reasons, including: an environment with such high stochasticity that it does not allow the algorithm to consistently find good solutions, an action space where the majority of actions do not yield good results, forcing the algorithms to need guidance to move towards sections of the space with better returns, the possibility that the algorithms used may not be sophisticated enough to solve such complex problems, and the absence of individual feedback on each firebreak, which translates into a sparse reward structure inherent to the problem. Secondly, the performance of the algorithm is anchored to that of the demonstrator. This is a double-edged sword: on one hand, it allows the agent to quickly reach a reasonable level of performance, but on the other, it causes the agent not to deviate much from the demonstrator, and if the demonstrator is far from optimal behavior, so will the algorithm. Thus, learning by demonstration as used in this work is recommended if 1) a heuristic is known that solves the problem quickly, and 2) the performance of this heuristic is not too far from the optimal behavior (or the one that is sought to be achieved).

A second motivation for this work was to see if these techniques could be used to improve the performance of a sub optimal algorithm. In this case, the algorithm chosen to be improved was the DPV, currently the best to solve the FPP. This was fully accomplished, surpassing this algorithm in all instances, with some showing a greater difference than others, but in all cases nonetheless. It is worth highlighting that even though our results surpass the DPV based heuristic, this last is considerably faster and able to solve much larger instances.

A notable element of using RL techniques in scenarios like the one above is the ease with which problem-specific constraints can be incorporated. For this, it is sufficient to restrict the agent's behavior at the level of the neural network, in contrast to what it means to do so in a mixed-integer programming model. Setting constraints in these latter models is generally complex and can create complications in the associated resolution algorithms.

In contrasting our research with those of \cite{lauer2017spatial} and \cite{large-scale_rl}, we discern a range of approaches and objectives within forest management paradigms. Lauer's research is anchored in the economic incentives derived from timber production, utilizing stochastic models and heuristic optimization to maximize the net present value of harvested timber. Altamimi, conversely, aims to maximize the total volume of stands, employing reinforcement learning algorithms such as DQN and A2C to determine treatments for each stand, thereby influencing fire probability. Our research, in contrast, is geared towards minimizing the environmental impact associated with biomass burning. Also, while Altamimi's employs Feedforward Neural Networks (FFNNs) as approximating functions within their RL algorithms, we use Convolutional Neural Networks (CNNs) within our RL algorithms to view the forest landscape as a multi-dimensional image. This approach leverages the spatial context and multivariate attributes of the landscape, thereby enhancing the precision and effectiveness of our fire management strategies.

While \cite{lauer2017spatial} focuses on economic optimization and \cite{large-scale_rl} on volume maximization, our study emphasizes the importance of sustainability and ecological conservation. This reflects a methodological advancement by integrating artificial intelligence technologies to more effectively address current ecological challenges. Our approach signifies a shift from financial or volume-centric traditional objectives towards a model that prioritizes environmental preservation, demonstrating the evolving landscape of forest management strategies in response to the pressing demands of ecological stewardship.

In conclusion, this research significantly advances the field by offering a dual contribution. Firstly, it introduces the innovative use of RL techniques as a novel solution to the FPP, thereby distinguishing itself from traditional OR methods and establishing RL as a promising alternative. Secondly, through a thorough comparison of different algorithms, this study validates their effectiveness and suitability for the specific challenge of firebreak placement. To the best of our knowledge, this study marks the first application of this approach to the problem at hand, signifying an important step in exploring alternative methods for tackling complex operational challenges in wildfire management. 

We identify several avenues to further advance the proposed methodology in future work. The first point considers the \textit{problem's structure}. The algorithms developed herein are each tailored to solve a specific instance, and as such, their ability to generalize across various instances falls beyond the scope of this work. Furthermore, it is a widely known fact that the before-mentioned sparse reward structure is detrimental to learning. One possible avenue for exploration could be to consider the historical performance of a specific firebreak each time a fire simulation occurs. 

Secondly, we identify the potential of \textit{more advanced algorithms}. Since the focus of our study is on effective firebreak placement, we did not use a category of more sophisticated algorithms known as Policy Gradient (PG). Within this group, certain algorithms are regarded as state-of-the-art, including Trust Policy Optimization \citep{trustpolicy} and Proximal Policy Optimization \citep{ppo}. The rationale for their exclusion lies in the complexities associated with incorporating demonstrations into their learning processes.

Finally, an \textit{alternative modeling strategy} merits consideration: the adoption of pre-defined connected patterns of firebreaks. Some research has been conducted on the optimal shapes for grouping firebreaks, such as the ones used in \citep{paper_david}. Implementing these u-shaped groups could significantly reduce the action space, focusing primarily on determining the central placement and orientation of each figure. This approach, through simplifying the decision-making process, could potentially broaden the scope and applicability of this research.

%\printcredits

\section*{Declaration of competing interest}
The authors declare that they have no known competing financial interests or personal relationships that could have appeared to influence the work reported in this paper.

\section*{Acknowledgments}

\textbf{JC} acknowledges the support of the Agencia Nacional de Investigaci\'on y Desarrollo (ANID), Chile, through funding Postdoctoral Fondecyt project No 3210311. This project has received funding from the European Union’s Horizon 2020 research and innovation programme under grant agreement No 101037419.

\bibliographystyle{cas-model2-names}

\bibliography{cas-refs}
\newpage
\appendix
\section{Appendices.}
\label{S:Apendix}
\begin{figure}[pos=h]
    \centering
    \includegraphics[width=0.5\textwidth]{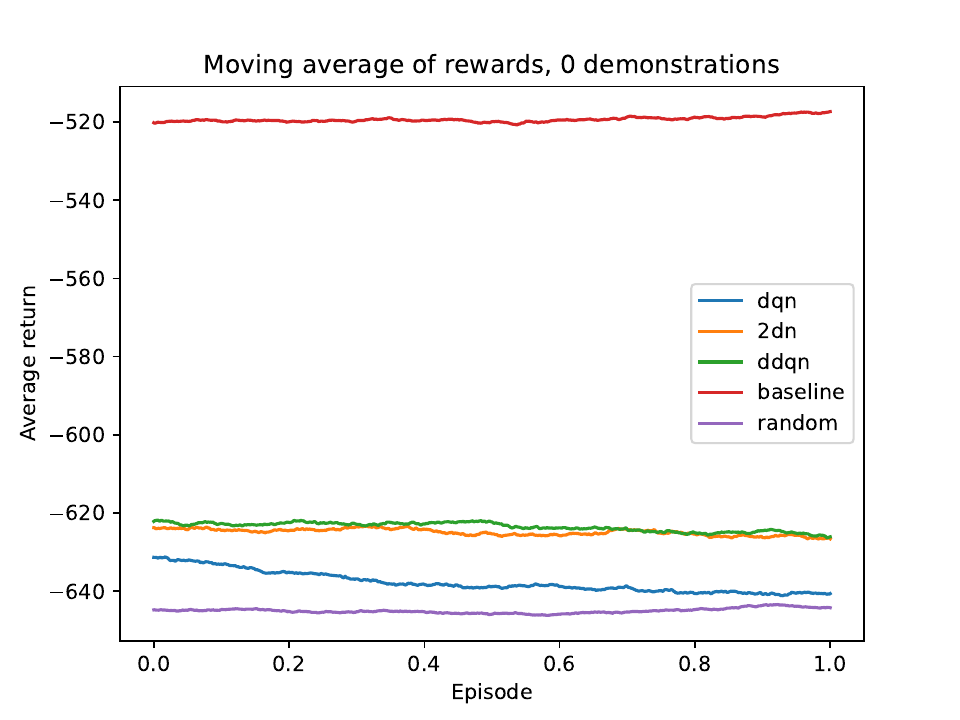}
    \caption{Results obtained without the incorporation of learning from demonstration using efficient-net in \texttt{Sub20}.}
    \label{fig:no-demonstrations}
\end{figure}

\begin{table}[pos=h]
    \centering
    
    \begin{tabular}{|l|l|l|}
        \hline
           & small-net   & big-net\\ \hline
        \textbf{D}         & 100000       & 100000               \\ \hline
        \textbf{BS}          & 64         & 64                \\ \hline
        \textbf{C}       & 200            & 200             \\ \hline
        \textbf{epochs}       & 20000     & 20000                  \\ \hline
        $\boldsymbol \alpha$        & $5e^{-5}$ & $5e^{-4}$ \ \\ \hline
        $\boldsymbol \gamma$               & 1      & 1                    \\ \hline
        $\boldsymbol \varepsilon$             & 1       & 1                   \\ \hline
        $\boldsymbol{\overline{\varepsilon}}$ & 0.005      & 0.005                \\ \hline
        $\boldsymbol{\hat{\varepsilon}}$             & 0.001    & 0.001                  \\ \hline
    \end{tabular}%
    \caption{D is the memory buffer size, BS batch size, C the target network's update frequency, $\overline{\varepsilon}$ the decay and $\hat{\varepsilon}$ the minimum value for $\varepsilon$.}
    \label{parameter-table}
\end{table}

\clearpage

\end{document}